\documentclass{ieeeaccess}
\usepackage{cite}
\usepackage{amsmath,amssymb,amsfonts}
\usepackage{algorithmic}
\usepackage{graphicx}
\usepackage{textcomp}
\usepackage{graphicx}
\usepackage{hyperref}
\usepackage{booktabs}
\usepackage{comment}
\usepackage{multirow}
\usepackage{multicol}
\usepackage{booktabs}
\usepackage{pifont}

\def\BibTeX{{\rm B\kern-.05em{\sc i\kern-.025em b}\kern-.08em
    T\kern-.1667em\lower.7ex\hbox{E}\kern-.125emX}}

\usepackage{graphicx}
\usepackage[font={sf,scriptsize},           
        labelfont={bf,color=accessblue},
        caption=false]                  
       {subfig} 
\begin{document}
\history{Paper under review}
\doi{TBD.}

\title{Cityscape-Adverse: Benchmarking Robustness of Semantic Segmentation with Realistic Scene Modifications via Diffusion-Based Image Editing} 
\author{\uppercase{Naufal Suryanto}\authorrefmark{1}, 
\uppercase{Andro Aprila Adiputra \authorrefmark{2},
Ahmada Yusril Kadiptya \authorrefmark{2},
Thi-Thu-Huong Le \authorrefmark{3},
Derry Pratama \authorrefmark{2},
Yongsu Kim \authorrefmark{4},
and Howon Kim}.\authorrefmark{2,4},
}
\address[1]{IoT Research Center, Pusan National University, Busan 46241, Republic of Korea (email: naufalsuryanto@gmail.com)}
\address[2]{School of Computer Science and Engineering, Pusan National University, Busan 46241, Republic of Korea (email: \{andro, yusril\}@pusan.ac.kr; derryprata@gmail.com).}
\address[3]{Blockchain Platform Research Center, Pusan National University, Busan 46241, Republic of Korea (email: lehuong7885@gmail.com).}
\address[4]{SmartM2M, Busan 46300, Republic of Korea (email: yongsu@smartm2m.com). }
\tfootnote{This research was supported by the MSIT(Ministry of Science and ICT), Korea, under the Convergence security core talent training business(Pusan National University) support program(RS-2022-II221201) supervised by the IITP(Institute for Information \& Communications Technology Planning \& Evaluation)}

\markboth
{N. Suryanto \headeretal: Cityscape-Adverse: Benchmarking Robustness of Semantic Segmentation via Diffusion-Based Image Editing}
{N. Suryanto \headeretal:  Cityscape-Adverse: Benchmarking Robustness of Semantic Segmentation via Diffusion-Based Image Editing}

\corresp{Corresponding author: Howon Kim (e-mail: howonkim@pusan.ac.kr).}

\begin{abstract}
    Recent advancements in generative AI, particularly diffusion-based image editing, have enabled the transformation of images into highly realistic scenes using only text instructions. This technology offers significant potential for generating diverse synthetic datasets to evaluate model robustness. In this paper, we introduce Cityscape-Adverse, a benchmark that employs diffusion-based image editing to simulate eight adverse conditions, including variations in weather, lighting, and seasons, while preserving the original semantic labels. We evaluate the reliability of diffusion-based models in generating realistic scene modifications and assess the performance of state-of-the-art CNN and Transformer-based semantic segmentation models under these challenging conditions. Additionally, we analyze which modifications have the greatest impact on model performance and explore how training on synthetic datasets can improve robustness in real-world adverse scenarios. Our results demonstrate that all tested models, particularly CNN-based architectures, experienced significant performance degradation under extreme conditions, while Transformer-based models exhibited greater resilience. We verify that models trained on Cityscape-Adverse show significantly enhanced resilience when applied to unseen domains. Code and datasets will be released at \url{https://github.com/naufalso/cityscape-adverse}.
\end{abstract}


\titlepgskip=-15pt

\maketitle

\section{Introduction}
\label{sec:introduction}

Semantic segmentation, the task of partitioning an image into distinct segments and classifying each segment into a predefined category, is pivotal in various real-world applications, particularly in autonomous driving \cite{ulku2022survey,muhammad2022vision}. Autonomous vehicles rely on accurate segmentation to understand their surroundings and make safe decisions, especially under different environmental conditions.

Despite recent advancements, semantic segmentation models often struggle to maintain accuracy in out-of-distribution (OOD) scenarios such as varying weather, lighting conditions, and seasonal changes \cite{hendrycks2021many,bruggemann2023refign}. For autonomous vehicles to operate reliably across diverse environments, segmentation models must be capable of generalizing effectively, a task that many current approaches still fail to meet under adverse conditions \cite{muhammad2022vision,zhang2023perception}.

Recent advancements in generative models, particularly diffusion-based image editing techniques, have shown promise in generating realistic image modifications \cite{brooks2023instructpix2pixlearningfollowimage,huang2024diffusion}. These generative models, used for data augmentation and robustness testing, have gained traction in various computer vision tasks such as object detection and image classification \cite{xu2023comprehensive, kumar2024image}. By enabling the creation of synthetic datasets that mimic real-world variations, these techniques provide a valuable resource for evaluating and enhancing the robustness of semantic segmentation models.

In this paper, we introduce \textbf{Cityscape-Adverse}, a novel benchmark designed to rigorously evaluate the robustness of semantic segmentation models. This benchmark utilizes diffusion-based image editing techniques to introduce realistic environmental variations to the well-known Cityscapes dataset \cite{cordts2016cityscapes}. By simulating conditions such as adverse weather or poor lighting, we aim to replicate OOD scenarios that are typically difficult and expensive to capture through traditional data collection methods \cite{zhou2017scene}. We hypothesize that these synthetic modifications provide an effective and scalable alternative to manually gathering and annotating large datasets in challenging environments. This approach reflects the increasing reliance on synthetic data in machine learning, particularly in areas where real-world data acquisition is costly or impractical \cite{nikolenko2021synthetic}.

The main contributions of this paper can be summarized as follows:
\begin{itemize}

\item \textbf{Introduction of the Cityscape-Adverse Benchmark}: We introduce \textbf{Cityscape-Adverse}, a novel benchmark designed to evaluate the robustness of semantic segmentation models under a wide range of simulated adverse environmental conditions. Unlike traditional datasets, our benchmark incorporates diverse and realistic conditions, providing a more comprehensive and rigorous framework for assessing model performance.

\item \textbf{Leveraging of Generative AI for Robust Semantic Segmentation}: This work investigates the application of state-of-the-art diffusion-based image editing to semantic segmentation. By utilizing advanced generative AI model, we demonstrate how these techniques enable mask-free image editing, preserving original semantic labels while conveniently generating datasets that reflect various environmental conditions. This approach offers substantial potential for enhancing model robustness and performance evaluation.

\item \textbf{Creation of Diverse Environmental Condition Datasets}: We developed 8 distinct datasets capturing various environmental conditions, including seasonal transitions (spring, autumn, winter), weather changes (rain, fog), and lighting variations (sunny, night, dawn). These diverse datasets facilitate a comprehensive evaluation of segmentation models, offering valuable insights into their behavior in dynamic, real-world scenarios.

\item \textbf{Out-of-Distribution (OOD) Robustness Evaluation}: We propose a novel framework for assessing and improving the robustness of semantic segmentation models in OOD scenarios by modifying real-world datasets. Through our Cityscape-Adverse benchmark, we demonstrate that synthetic datasets significantly enhance model robustness in real-world OOD environments, providing critical insights into the strengths and weaknesses of current models.

\item \textbf{Comprehensive Benchmarking of State-of-the-Art Models}: We conduct a comparative evaluation of leading semantic segmentation models using the Cityscape-Adverse benchmark. Our analysis highlights key strengths and limitations of existing models when faced with adverse conditions, offering practical guidance for future model development and training strategies.

\end{itemize}

\section{Related Work} 
\label{sec:related_work}

\subsection{Semantic Segmentation Models} 
Semantic segmentation is an essential task in practical Computer Vision. It aims to classify image pixels into predefined categories to provide a better understanding of visual scenes. Recently, significant advancements have been achieved using deep learning techniques, particularly Convolutional Neural Networks (CNNs) and Transformer-based models.

CNNs have been foundational for semantic segmentation since the early stages. ICNet \cite{zhao2018icnet} introduced an image cascade network that downsamples input images and fuses feature maps at multiple resolutions, achieving a balance between speed and accuracy. Similarly, DeepLabv3+ \cite{chen2018encoder} enhances performance by refining object boundaries, utilizing atrous separable convolutions in its encoder to capture multi-scale features, which are then processed by a decoder. This approach improves semantic understanding, especially in complex scenes. On the other hand, DDRNet \cite{hong2022deep} achieves state-of-the-art performance by combining high-resolution and low-resolution branches with Aggregation Pyramid Pooling, allowing it to capture rich contextual information with only a minor tradeoff in computational efficiency.

More recently, Transformer-based models have emerged, demonstrating superior scalability and effectiveness in capturing long-range dependencies. SETR \cite{zheng2021rethinking} revolutionizes the segmentation task by employing a pure Transformer encoder, treating the problem as sequence-to-sequence prediction and achieving excellent results by modeling the entire image as patches. SegFormer \cite{xie2021segformer} adopts a hierarchical Transformer design without positional encoding and pairs it with a simple Multi-Layer Perceptron decoder, leading to both efficiency and robustness. Mask2Former \cite{cheng2022mask2former} takes advantage of multi-scale attention mechanisms, replacing traditional cross-attention with masked attention to handle small objects more effectively and reduce computation costs without sacrificing performance.

In this study, we evaluate the robustness of various CNN and Transformer-based models—ICNet, DeepLabv3+, DDRNet, SETR, SegFormer, and Mask2Former—under adverse conditions in our benchmark.

\subsection{Scene Understanding Datasets for Urban Driving}
Urban driving scenarios present significant challenges for semantic segmentation due to the diversity and complexity of scenes. Several large-scale datasets have been developed to advance scene understanding in these contexts.

The Cityscapes \cite{cordts2016cityscapes} is a pioneering dataset, providing high-quality pixel-level annotations for 5,000 images captured across 50 cities in urban environments. It offers diverse street scenes with fine-grained semantic segmentation labels. The KITTI Vision Benchmark \cite{10.5555/2354409.2354978} supplies extensive data for various computer vision tasks, including semantic segmentation, with a focus on road and lane detection in urban and highway settings. Similarly, ApolloScape \cite{Huang_2020} is a large-scale dataset designed for autonomous driving, offering annotations for tasks such as semantic segmentation, instance segmentation, and 3D object detection.

For a broader range of conditions, BDD100K (Berkeley DeepDrive) \cite{yu2020bdd100k} includes 100,000 images with annotations for tasks such as object detection, lane detection, and segmentation. It captures diverse conditions, including weather and time-of-day variations, making it valuable for evaluating model robustness. The Mapillary Vistas dataset \cite{Neuhold2017TheMV} provides a global perspective with road scenes from various countries and fine-grained annotations across numerous object categories. Its updated version, Mapillary Vistas 2.0 \cite{vistasv2}, increases label granularity, featuring 124 semantic categories (70 instance-specifically annotated) compared to the 66 classes in the original version (37 instance-specific).

For low-light driving scenarios, the Nighttime Driving dataset \cite{dai2018dark} includes 35,000 road scene images captured from daytime to nighttime, with 50 finely annotated nighttime images, using the 19 semantic classes from Cityscapes. Dark Zurich \cite{SDV20} offers 8,779 images captured at nighttime, twilight, and daytime, with GPS coordinates provided for each image to match nighttime or twilight images with their daytime counterparts.

While these datasets have significantly advanced the development and evaluation of semantic segmentation models, they often lack the extensive variations in adverse conditions necessary for robust testing. Cityscape-Adverse addresses this gap by providing datasets that incorporate realistic environmental condition changes, allowing for a more comprehensive evaluation of model performance in challenging scenarios.

\subsection{Robustness in Out-of-Distribution Evaluation Methods}

Evaluating model robustness in out-of-distribution (OOD) scenarios is essential for understanding how models perform under diverse and unexpected conditions. This has become a growing area of research in computer vision and machine learning \cite{9782500, yang_generalized_2024, salehi2022unifiedsurveyanomalynovelty}.

Robustness evaluation ideally relies on real-world datasets that capture various adverse conditions. A prominent example is the ACDC (Adverse Conditions Dataset with Correspondences) \cite{sakaridis_acdc_2021}, which includes images taken under challenging conditions like nighttime, fog, rain, and snow. These datasets are valuable benchmarks for assessing the performance of semantic segmentation models under adverse weather and lighting scenarios.

Recent advancements in OOD evaluation have introduced more comprehensive benchmarks. OOD-CV \cite{zhao2022oodcv} and its extended version OOD-CV-v2 \cite{zhao2024oodcvv2} provide fine-grained benchmarks that evaluate model robustness to specific nuisances in natural images. These benchmarks span tasks like image classification, object detection, and 3D pose estimation, offering detailed analyses of model performance across various distribution shifts.

As the field evolves, there is a growing focus on developing more realistic and challenging benchmarks for OOD detection \cite{hendrycks2021facesrobustnesscriticalanalysis}. Researchers are also working on standardized evaluation protocols and metrics to provide a more comprehensive assessment of model robustness in real-world scenarios \cite{yang2021semanticallycoherentoutofdistributiondetection, hendrycks2019deepanomalydetectionoutlier}.

In addition to real-world datasets, synthetic data modifications offer a scalable approach for simulating OOD conditions. For example, the Foggy Cityscapes dataset \cite{sakaridis_semantic_2018} alters Cityscapes images to simulate fog, enabling controlled experiments that are difficult to achieve with real-world data \cite{michaelis2020benchmarkingrobustnessobjectdetection, hendrycks2019benchmarking}.

Recent advancements in generative AI have made synthetic image modification even more practical. The TSIT framework \cite{jiang2020tsit} utilizes a GAN-based image-to-image translation approach to generate realistic domain-transferred images using content images, style targets, and masks. This method is particularly valuable for augmenting datasets where labeled data is scarce, producing synthetic data that retains essential features for segmentation tasks.

The emergence of diffusion-based models has further advanced image generation, enabling the synthesis of high-quality images from text prompts \cite{rombach2022high}. Recent work has evaluated the reliability of these models in generating synthetic OOD data by fine-tuning ControlNet \cite{zhang2023adding} to preserve semantic consistency using segmentation masks \cite{loiseau2024reliability}.

Table \ref{tab:synthetic_dataset_comparison} compares our method with previous approaches. Our method uses a state-of-the-art pretrained diffusion-based image editing model, requiring only a source image and text prompt to perform scene modifications while preserving the original layout. Unlike other methods, our approach relies on the model's zero-shot capability and does not require retraining to perform these modifications.

\begin{table}
\centering
\caption{Comparison of out-of-distribution evaluation using Generative AI-based synthetic modification}
\begin{tabular}{|l|c|c|c|c|}
\hline
\textbf{Methods}                                         & \textbf{Retraining?}                & \textbf{Input}                   & \textbf{Base Model}                                                                      \\ \hline
TSIT \cite{jiang2020tsit}                                          & required         & \ding{117}, \ding{71}, \ding{60} & GAN                                                                                      \\ \hline
GenVal \cite{loiseau2024reliability}                               & required         & \ding{117}, \ding{60}, \ding{81} & \begin{tabular}[c]{@{}c@{}}Txt2Img Diffusion\\ (SD 1.5, SDXL)\\+ ControlNet\end{tabular} \\ \hline
\begin{tabular}[c]{@{}l@{}}Cityscape-Adverse \\(Ours)\end{tabular} & not required     & \ding{117}, \ding{81}            & \begin{tabular}[c]{@{}c@{}}Img2Img Diffusion\\ (CosXLEdit)\end{tabular}                  \\ \hline
\end{tabular}
\subfloat{Notes: \ding{117} source/content image, \ding{71} target image style, \ding{60} mask, \ding{81} prompt}
\label{tab:synthetic_dataset_comparison}
\end{table}

\subsection{Diffusion-Based Image Editing}

Diffusion in image generation models refers to an iterative process that refines images by simulating a denoising process. This technique has become a powerful tool for generating high-quality images and significantly advancing the field of generative AI \cite{NEURIPS2020_4c5bcfec}. One prominent application is Stable Diffusion, which employs a latent diffusion process with text-based controls to generate images from textual descriptions \cite{rombach2022high}. An upgraded version, Stable Diffusion XL (SDXL), achieves higher quality and resolution through a larger model and improved conditioning techniques \cite{podell2024sdxl}. This model has further evolved into Cos Stable Diffusion XL (CosXL), which incorporates a Cosine-Continuous EDM VPred schedule, enhancing its ability to produce a full-color range from deep blacks to bright whites and providing finer control over image adjustments at each refinement step \cite{stabilityai_cosxl, lin2024common}.

In diffusion-based image editing, a common approach is inpainting \cite{lugmayr2022repaint}, where specific regions of an image are edited using a mask. The diffusion model regenerates the masked area based on the context of its surroundings and, optionally, a guiding text prompt. This technique allows precise editing of image regions, adding considerable flexibility to the editing process. However, inpainting can only directly control spatial modifications. To address this, ControlNet was introduced, allowing the addition of various input conditions—such as sketches, normal maps, depth maps, poses, and semantic segmentation masks—to guide the diffusion model’s output. This method preserves the original diffusion model weights and uses a supplementary ControlNet module to steer the model output, expanding diffusion model controllability \cite{zhang2023adding}.

Another approach to diffusion-based image editing is introduced by InstructPix2Pix, which trains the model to perform edits directly from textual instructions without requiring additional input conditions \cite{brooks2023instructpix2pixlearningfollowimage}. This instruction-based editing method simplifies the editing process, making it more accessible for users. Our approach builds on this model by utilizing CosXLEdit, a variant of CosXL designed to handle similar instruction-based edits with enhanced fidelity.

\Figure[t!]()[width=\linewidth]{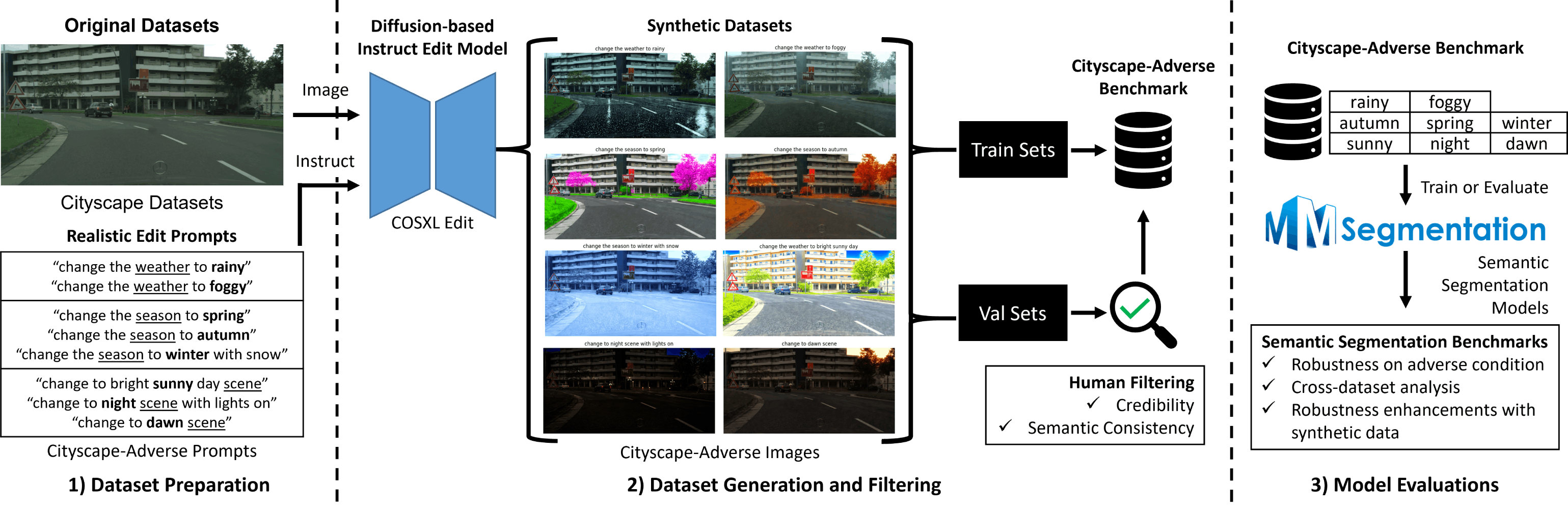}
{Our pipeline for generating and evaluating Cityscape-Adverse using diffusion-based instructed image editing.
The process includes dataset preparation, generation, human filtering, and benchmarking. \label{fig:cityscape-adverse-pipeline}}

\section{Cityscape-Adverse}
\label{sec:methodology}
This section describes our pipeline for generating the Cityscape-Adverse dataset and performing benchmarks, as illustrated in Fig. \ref{fig:cityscape-adverse-pipeline}.

\subsection{Dataset Preparation}
Our dataset is derived from the original Cityscapes dataset, utilizing 2,975 training images and 500 validation images. All evaluations in this paper are based on the validation set, as the ground-truth labels for the test set are not publicly released. We retain the original resolution of the images at $2048 \times 1024$ pixels for all transformations to ensure consistency and compatibility with previous benchmarks.

The primary goal of our modifications is to maintain semantic consistency, meaning the ground-truth annotations provided with Cityscapes remain valid after the transformations. This ensures the synthetic images are suitable for semantic segmentation tasks without needing new annotations.

The modifications are designed to reflect realistic changes in environmental conditions, categorized into three groups:
\begin{itemize}
    \item \textbf{Season changes:} Spring, autumn, and winter (snowy).
    \item \textbf{Weather changes:} Rainy and foggy.
    \item \textbf{Lighting conditions:} Sunny, night, and dawn.
\end{itemize}
This results in eight distinct dataset variants, each representing a different adverse condition that models must handle. For each environmental condition, we experimented with different prompts to guide the diffusion model, such as \textit{"change the weather to rainy"}, \textit{"change the season to winter with snow"}, and \textit{"change to night scene with lights on."} To enhance the realism of the generated images, we appended \textit{", photo-realistic"} to the prompts and adjusted the guidance scales (ranging from 5 to 7) to balance the fidelity of the environmental changes without altering the scene's semantic structure.

\subsection{Dataset Generation and Filtering}
To generate the synthetic images, we applied the pre-defined prompts to both the training and validation sets of the Cityscapes dataset. This generated a total of $2975 \times 8$ images for the training set and $500 \times 8$ images for the validation set, across the eight different environmental conditions. The resulting synthetic images form the Cityscape-Adverse benchmark, designed to test the robustness of semantic segmentation models in adverse scenarios.

\subsubsection{Generation Process}
For generating the images, we used the CosXLEdit model \cite{stabilityai_cosxl}, which is optimized for instruction-based image editing. This model was selected for its ability to generate high-quality, realistic images with fewer steps due to features like v-prediction, ZeroSNR, and a cosine schedule \cite{lin2024common}. These features were particularly effective in producing modifications for extreme lighting conditions, such as generating bright sunny scenes or realistic night-time images, which are challenging for many generative models.

The image generation process was automatic, applying the diffusion model based on the provided prompts. However, the quality and suitability of these synthetic images for use in a semantic segmentation benchmark were not guaranteed by the model alone. Therefore, additional steps were taken to ensure the credibility and consistency of the dataset.

\subsubsection{Human Filtering for Quality Control}
\label{subsubsec:human_filtering}
After generating the synthetic images, we conducted a manual verification process to ensure they met two criteria:
\begin{itemize}
    \item \textbf{Credibility:} The images should realistically represent the intended environmental conditions, such as rain or snow. For instance, a scene modified to appear rainy should contain realistic raindrops, wet surfaces, and appropriate atmospheric effects, without obvious artifacts.
    \item \textbf{Semantic Consistency:} The modifications should not alter the spatial layout or introduce new objects that would affect the ground-truth segmentation labels. For example, while changing a daytime scene to nighttime, objects like cars, trees, and buildings should remain unaltered in position and shape, even though the lighting conditions change.
\end{itemize}

We conducted human filtering primarily on the validation set to ensure it remained a credible benchmark for evaluating model performance. Given the importance of the validation data in benchmarking, special attention was paid to ensuring that the semantic consistency of these images was preserved. Any images where the scene structure was altered by the diffusion process were discarded.

The criteria for acceptance during human filtering emphasized \textbf{semantic consistency} over complete realism. While some images might not have achieved perfect realism (e.g., minor color distortions), they were retained if the core objects in the scene remained easily recognizable by a human observer. This decision reflects our goal of creating a challenging yet practical benchmark for evaluating model robustness under extreme conditions. The final acceptance rates and the resulting dataset sizes are presented in Table \ref{table:acceptance_data}.

\begin{table}[t]
\centering
\caption{Total filtered validation data used in our Cityscape-Adverse.}
\begin{tabular}{lcccc}
\toprule
& \textbf{Rainy} & \textbf{Foggy} & \textbf{Spring} & \textbf{Autumn}  \\
\midrule
\textbf{Total Accepted} & 495     & 478    & 493     & 466  \\
\textbf{Acceptance Rate} & 99.0\% & 95.6\% & 98.6\%  & 93.2\% \\
\bottomrule
& \textbf{Snow} &\textbf{Sunny} & \textbf{Night} & \textbf{Dawn} \\
\midrule
\textbf{Total Accepted} & 488 & 500 & 500 & 500  \\
\textbf{Acceptance Rate} & 97.6\%  & 100.0\% & 100.0\% & 100.0\%  \\
\bottomrule
\end{tabular}
\label{table:acceptance_data}
\end{table}

\subsubsection{Training Set Considerations}
Unlike the validation set, we did not apply manual filtering to the training data. Based on the high acceptance rate of the validation images, we hypothesized that the synthetic images generated from the training set were of sufficient quality to be used directly in training. Additionally, skipping human filtering for the training set allows us to test whether models trained on purely synthetic data—without any manual intervention—can still achieve high robustness.

This approach also allows us to test the feasibility of using synthetic data generated by CosXLEdit for training, which is critical for scalability. By not filtering the training set, we aim to evaluate whether models trained on synthetic data alone can achieve robustness comparable to those trained on real-world data. A detailed qualitative evaluation of the generated images, comparisons with another state-of-the-art image editing model, and examples of failure cases are presented in Sec. \ref{subsec:feasibility-studies}.

\subsection{Model Evaluation}
We organized the generated data, including the training and filtered validation sets, into the Cityscape-Adverse benchmark. Each dataset reflects a specific environmental condition (e.g., spring, autumn, rainy, night), while retaining the original ground-truth annotations from Cityscapes to ensure consistency in semantic labeling.

For our evaluations, we used the MMSegmentation framework \cite{mmsegmentation}, a PyTorch-based toolbox that offers a wide range of pre-trained semantic segmentation models. The framework's support for both training and testing facilitated efficient benchmarking across various models and datasets.

Our evaluation process encompasses three key analyses:

\begin{itemize}
    \item \textbf{Robustness Evaluation on Adverse Conditions:} We evaluated models on the Cityscape-Adverse datasets to assess their performance under different environmental conditions using mean Intersection over Union (mIoU) as a widely used metric in semantic segmentation. This analysis highlights the resilience of baseline models to weather, season, and lighting changes.

    \item \textbf{Cross-Dataset Analysis:} We analyzed the generalization capabilities of models trained on specific subsets of Cityscape-Adverse by evaluating them on other datasets within the benchmark. This analysis highlights correlations between different environmental conditions and assesses how well models can generalize across diverse modifications.

    \item \textbf{Robustness Enhancements with Synthetic Data:} We evaluated the robustness of models trained on Cityscape-Adverse by testing them on multiple real-world adverse datasets with compatible semantic labels. Additionally, we compared the effectiveness of our synthetic datasets with previous synthetic augmentation methods from the literature.
\end{itemize}

\section{Experiments and Results}
\label{sec:experiments_and_result}
This section presents the main experiments conducted for the Cityscape-Adverse benchmark. We begin by describing the experimental setup, followed by an initial evaluation of how well the Diffusion-based Instruct Edit model generates synthetic datasets that meet our predefined criteria. Next, we provide a robustness evaluation of existing semantic segmentation models on the Cityscape-Adverse benchmark, analyzing the influence of each dataset type on model performance. Finally, we demonstrate how synthetic datasets can enhance model robustness when applied to real-world datasets.

\subsection{Experimental Setup}
\label{subsec:experiments}

\subsubsection{Datasets}
We used Cityscapes \cite{cordts2016cityscapes} as the baseline dataset for all our experiments. To simulate various adverse conditions, each dataset was transformed into a Cityscape-Adverse dataset, as detailed in Section \ref{sec:methodology}.

To evaluate the enhancement of model robustness in real-world adverse scenarios, we selected several benchmark datasets with compatible semantic labels, including ACDC \cite{sakaridis_acdc_2021}, BDD100K \cite{yu2020bdd100k}, Dark Zurich \cite{SDV20}, and Night Driving \cite{dai2018dark}. Additionally, we incorporated ApolloScape \cite{Huang_2020} and Mapillary Vistas v2 \cite{vistasv2}, applying label conversion to match the label format of Cityscapes. All evaluation results in our experiments are based on the validation sets, as the ground truth for the test sets is not always publicly available.

\subsubsection{Compared Methods for Generating Synthetic Datasets}
We evaluated CosXLEdit \cite{stabilityai_cosxl} as our selected instruction-based image editing model and qualitatively compared it to the leading InstructPix2Pix model \cite{brooks2023instructpix2pixlearningfollowimage}.

To benchmark against previous methods, we included the recent GenVal approach \cite{loiseau2024reliability} and generated synthetic training data using the same modification types as Cityscape-Adverse to ensure a fair comparison with our proposal. We used their official code\footnote{\url{https://github.com/valeoai/GenVal}} and pretrained model to generate the modified data by adjusting the image prompts.


\subsubsection{Evaluated Semantic Segmentation Models}
We selected several state-of-the-art semantic segmentation models pre-trained on the ImageNet1k dataset \cite{imagenet}, available through the MMSegmentation framework, to serve as baseline models for our evaluation. The chosen models include  DeepLabV3+ \cite{chen2018encoder}, ICNet \cite{zhao2018icnet}, and DDRNet \cite{hong2022deep} as representatives of CNN-based architectures, and SegFormer \cite{xie2021segformer}, SETR \cite{zheng2021rethinking}, and Mask2Former \cite{cheng2022mask2former} as Transformer-based models. The detailed model specifications, including backbone architecture, crop size, memory usage, and inference time, are provided in Table \ref{tab:model_specs}, sourced from the official MMSegmentation repository \cite{mmsegmentation}.

\begin{table}[t]
\centering
\caption{Specifications of the evaluated models, including backbone architecture, crop size, memory usage, and inference time.}
\resizebox{\linewidth}{!}{
\begin{tabular}{|l|c|c|c|c|}
\hline
\textbf{Method}         & \textbf{Backbone} & \textbf{Crop Size} & \textbf{Mem (GB)} & \textbf{Infer (fps)} \\ \hline
DeepLabV3+               & R-50-D8           & $512 \times 1024$           & 7.5               & 3.94                 \\ \hline
ICNet                   & R-50-D8           & $832 \times 832$            & 2.53              & 20.08                \\ \hline
DDRNet                  & DDRNet23          & $1024 \times 1024$          & 7.26              & 33.41                \\ \hline
SegFormer                & MIT-B0            & $1024 \times 1024$          & 3.64              & 4.74                 \\ \hline
SETR Naive               & ViT-L             & $768 \times 768$            & 24.06             & 0.39                 \\ \hline
Mask2Former              & Swin-S            & $512 \times 1024$           & 8.09              & 5.57                 \\ \hline
\end{tabular}
}
\label{tab:model_specs}
\end{table}

\begin{figure*}
\centering
\subfloat[Cityscape-Adverse dataset generated by CosXLEdit]{\includegraphics[width=\columnwidth]{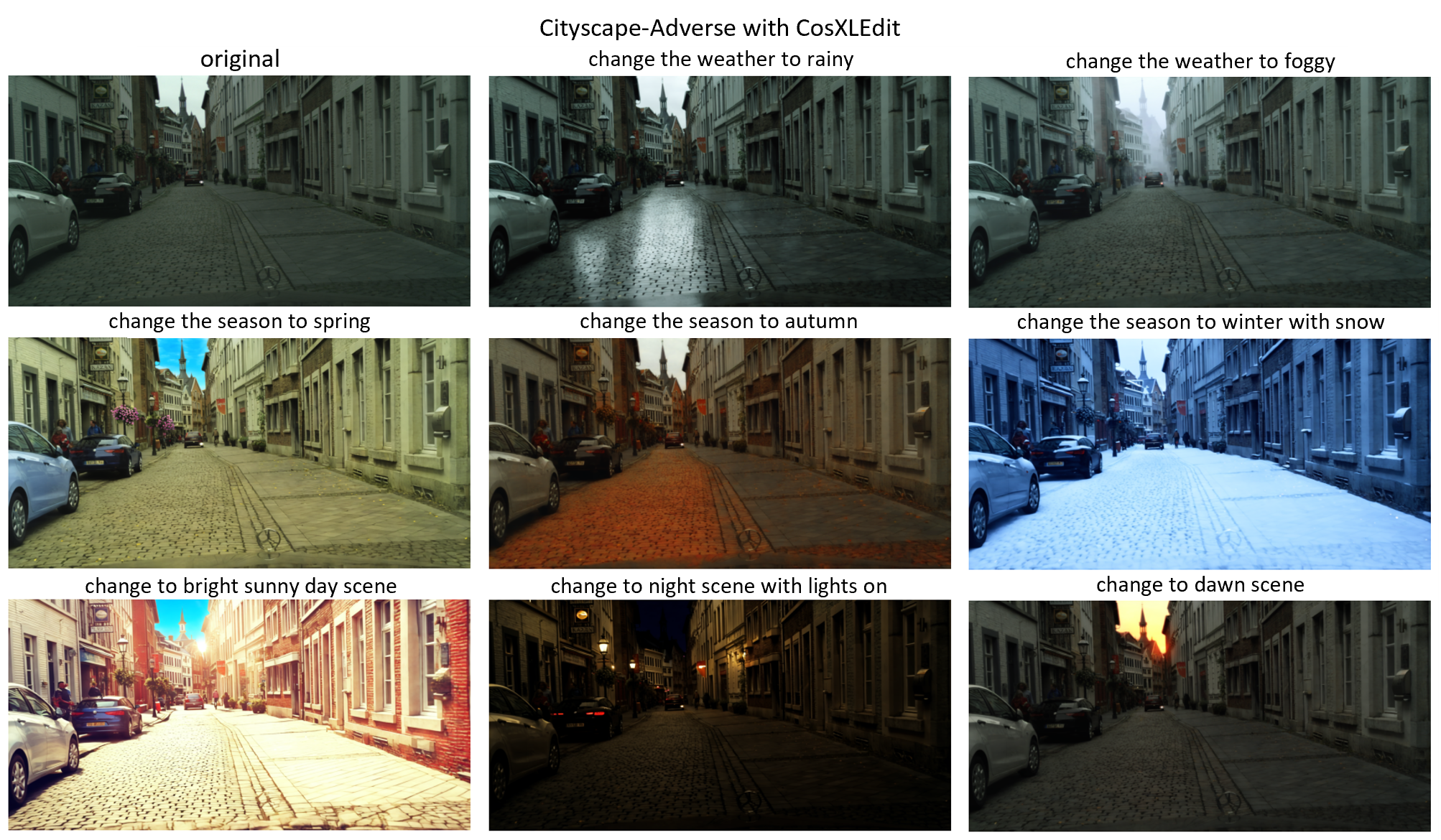}}
\hfil
\subfloat[Cityscape-Adverse dataset generated by InstructPix2Pix]{\includegraphics[width=\columnwidth]{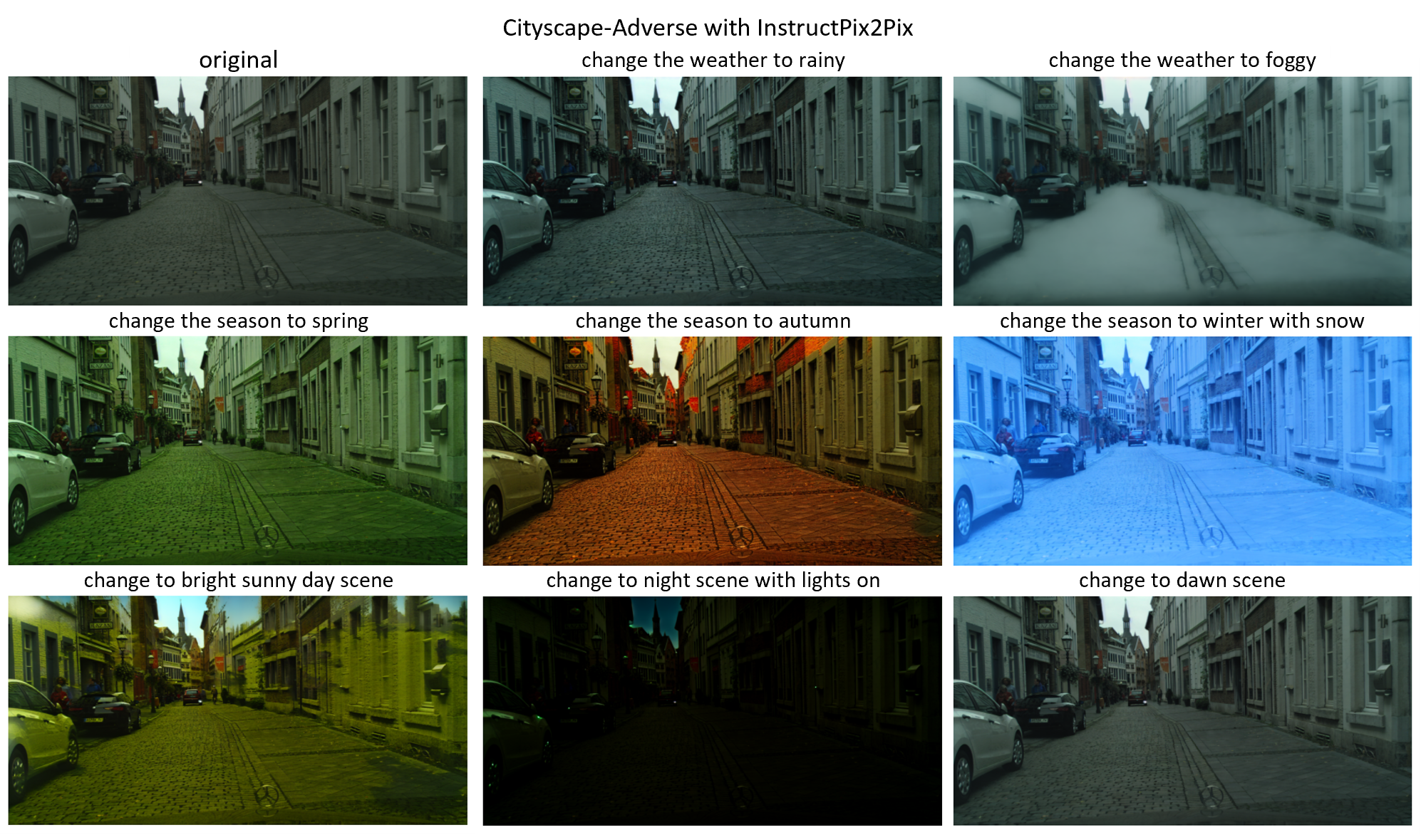}}
\caption{Qualitative comparison between CosXLEdit \cite{stabilityai_cosxl} and InstructPix2Pix \cite{brooks2023instructpix2pixlearningfollowimage} for modifying the original dataset in the adverse dataset setting. All prompts include the postfix ", photo-realistic" to enforce generation towards realism. Zoom in for the details.}
\label{fig:cosxl_vs_instructpix2pix}
\end{figure*}

\subsubsection{Evaluation Metrics}
We primarily evaluate model performance using Intersection over Union (IoU) and mean Intersection over Union (mIoU), which are standard metrics in semantic segmentation. The IoU for each class is calculated as the ratio of correctly predicted pixels to the union of ground-truth and predicted pixels for that class. It is defined as:

\begin{equation}
    \text{IoU} = \frac{TP}{TP + FP + FN}
\end{equation}

where $TP$, $FP$, and $FN$ represent the true positives, false positives, and false negatives, respectively, for a given class. The mIoU is the average of the IoUs across all classes, providing a single scalar value that summarizes the overall segmentation performance:

\begin{equation}
    \text{mIoU} = \frac{1}{C} \sum_{i=1}^{C} \text{IoU}_i
\end{equation}

where $C$ is the total number of classes, and $\text{IoU}_i$ is the IoU for class $i$.

In addition to mIoU, we measure both the performance drop and performance enhancement to capture how model performance changes when exposed to adverse conditions compared to the original Cityscapes dataset. The performance drop quantifies the degradation in segmentation performance, while the performance enhancement highlights improvements in adverse scenarios due to synthetic data augmentation.

The performance drop is calculated as the difference between the mIoU on the original dataset and the mIoU on the modified Cityscape-Adverse dataset:

\begin{equation}
    \text{Performance Drop} = \text{mIoU}_{\text{original}} - \text{mIoU}_{\text{adverse}}
\end{equation}

Similarly, the performance enhancement (Enh.) is defined as the increase in mIoU when comparing models trained with enhanced data to those trained on the original Cityscapes:

\begin{equation}
    \text{Performance Enh.} = \text{mIoU}_{\text{enhanced}} - \text{mIoU}_{\text{original}}
\end{equation}

These metrics provide insights into how well models generalize to various adverse environmental conditions, such as different weather, lighting, or seasonal changes, while also evaluating the benefits of synthetic data augmentation or other techniques in improving robustness.

\subsection{Feasibility Studies}
\label{subsec:feasibility-studies}
We carefully evaluate the feasibility of existing instruction-based image editing models and previous work methods to synthesize a modified dataset based on our defined criteria in Sec. \ref{subsubsec:human_filtering}. This step is crucial as it ensures the validity of the synthetic dataset before it is employed as a benchmark for the semantic segmentation.

\subsubsection{Comparison with Another Instruction-based Image Editing}
\label{subsubsection:other-instruct-edit-model}
As illustrated in Figure \ref{fig:cosxl_vs_instructpix2pix}, CosXLEdit consistently produces more realistic and detailed environmental transformations across various weather conditions, seasons, and lighting scenarios compared to InstructPix2Pix.

In weather transformations such as rain and fog, CosXLEdit excels in creating more immersive effects. Rain scenes feature realistic wet surface reflections and atmospheric dimming, while fog transformations introduce a natural gradient that progressively reduces visibility. In contrast, InstructPix2Pix produces flatter results, with rain lacking reflections and fog appearing more uniform and less dynamic, leading to less convincing weather effects.

For seasonal changes, CosXLEdit effectively modifies color palettes and lighting to represent spring, autumn, and winter. Spring scenes display vibrant greenery with natural lighting, autumn scenes are rich with warm, earthy tones, and winter captures realistic snow textures and cold hues. InstructPix2Pix, however, often applies more uniform color shifts—spring is overly green, and autumn has unnatural reddish hues. Winter transformations in InstructPix2Pix also lack snow detail, making these scenes appear less convincing.

In lighting transformations (sunny day, night, and dawn), CosXLEdit handles brightness, shadows, and light sources with greater accuracy. Sunny scenes feature well-defined shadows and realistic highlights, while night scenes include soft, natural streetlights. CosXLEdit’s dawn transformations show smooth lighting transitions. In contrast, InstructPix2Pix tends to oversaturate sunny scenes, while its night and dawn transformations show abrupt lighting changes, leading to less immersive results.

CosXLEdit’s superior performance is largely attributed to its use of the \textit{Cosine-Continuous Exponential Denoising Model (EDM) VPred schedule}, which expands the color spectrum and enhances the model’s adaptability for seamless transitions across iterations. InstructPix2Pix, relying on an older Stable Diffusion version, lacks these refinements. Both models preserve semantic consistency, ensuring object positions remain intact for segmentation tasks, though CosXLEdit achieves this with greater precision.

\subsubsection{Comparison with Previous Methods}

\Figure[]()[width=\linewidth]{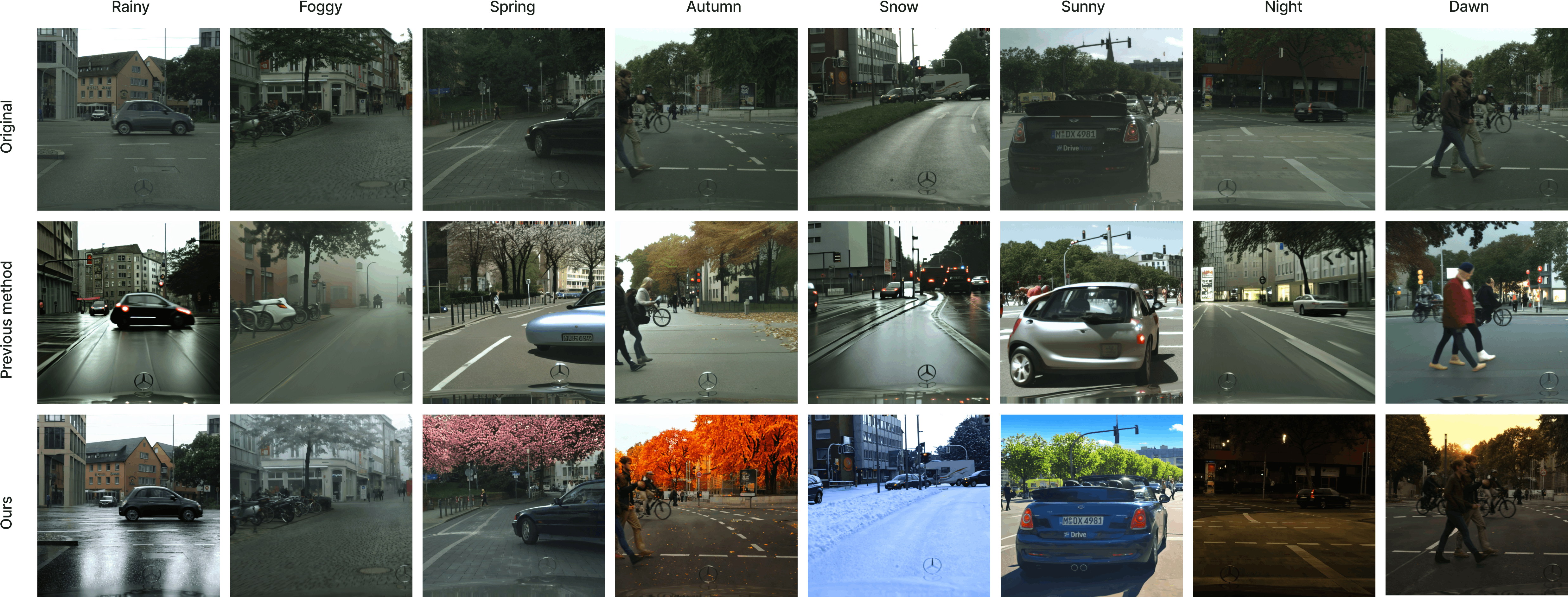}
{Examples of generated multiple scene dataset result from the previous method \cite{loiseau2024reliability} compared to ours. Zoom in for the details. \label{fig:dataset-quality-comparison}}

Figure~\ref{fig:dataset-quality-comparison} highlights the significant improvement in visual fidelity achieved by our method compared to the previous approach \cite{loiseau2024reliability} across various environmental conditions. For instance, in the autumn scene, the previous method produces degraded road markings, blurry leaves, and unrealistic pedestrian shapes. In contrast, our method generates sharper and more realistic results, preserving the road markings, rendering leaves crisply, and depicting pedestrians naturally, with the added effect of falling leaves. Although the leaf colors in our results are slightly more vibrant than typical autumn foliage, this does not detract from the overall realism of the image.

In challenging conditions, our approach consistently delivers vibrant colors while maintaining fine details. The previous method, however, often results in over-smoothed textures and washed-out colors. For example, in foggy conditions, our method preserves the atmospheric haze while maintaining the sharpness of object boundaries. Similarly, in dawn scenes, our approach captures the transformation of sunlight, realistically adjusting objects to appear backlit. A notable improvement is also observed in the night scene, where our method accurately adjusts lighting to depict nighttime conditions while preserving the original dataset's features.

The previous method frequently introduces unrealistic distortions in various scenarios, particularly in buildings, cars, and roads. A striking example occurs in the snowy scene, where road lanes are distorted, and the overall atmosphere fails to convey a snowy environment. In contrast, our method retains the integrity of road lanes while realistically adding snow piles to roads and grassy areas. Moreover, the previous method distorts the shape of cars in spring and sunny scenes, whereas our approach preserves the details of road lanes and renders vibrant green colors in the sunny scene, along with a realistic depiction of blooming sakura trees in spring.

The shortcomings of the previous method in maintaining fidelity and preserving original image content can be attributed to its approach. It attempts to extract image information as a text prompt and regenerate a modified image using a text-to-image diffusion model conditioned on a semantic segmentation mask via ControlNet \cite{Zhang_2023_ICCV}. However, this method can lose key attributes since text prompts may not fully capture the original image's details. In contrast, our method uses an image-to-image instruction-based diffusion model, trained specifically to modify the source image according to an instruction prompt. Additionally, the previous method relies on an older version of Stable Diffusion, which contributes to its lower-quality results.

\subsubsection{Our Failure Cases}

\Figure[]()[width=\linewidth]{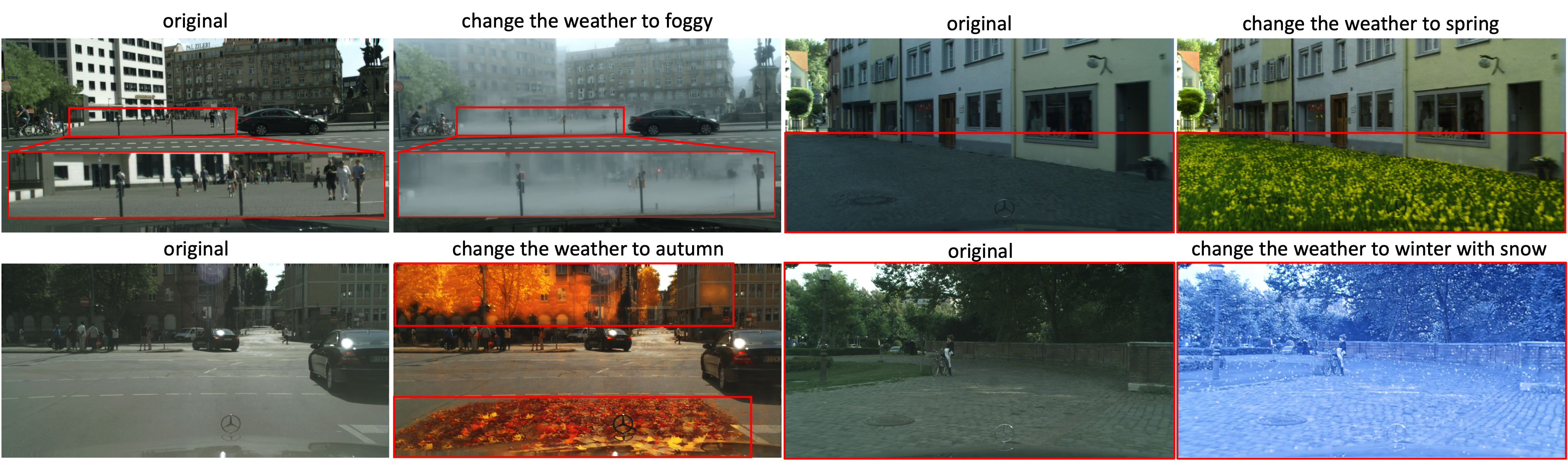}
{Examples of failure cases using CosXLEdit. The top row illustrates instances where semantic consistency is compromised, leading to altered ground-truth labels. For example, persons are missing in the foggy dataset, and the road in the spring dataset is replaced by a flower garden. The bottom row highlights credibility issues due to unrealistic modifications, such as unnatural leaves, poorly blended orange hues in the autumn scene, and excessive tinting in the winter modification. Zoom in for details. \label{fig:failure_case_sample}}

While CosXLEdit generally delivers strong performance, it does not always perfectly meet our criteria. As indicated in Table \ref{table:acceptance_data}, several examples from various weather conditions were rejected during the human filtering process for failing to meet these standards (see Figure \ref{fig:failure_case_sample} for the examples).

One common issue observed in \textit{rainy} modifications was the road surface appearing overly glossy, deviating from the look of a normally wet road. In \textit{foggy} scenarios, we encountered cases where parts of a person turned into mist while other parts remained clear, creating a jarring visual inconsistency. This not only impacted realism but also compromised semantic labels, which could be problematic for downstream tasks like segmentation.

In \textit{spring} transformations, we occasionally observed rare instances where the road was replaced by a flower-filled garden, altering its semantic label from walkable road to non-walkable vegetation, thus violating semantic consistency. \textit{Autumn} transformations exhibited the most failures, with unrealistic leaf textures and poorly blended orange hues. In some instances, new trees appeared on top of buildings or in the sky, altering original scene labels. Additionally, some people or objects turned orange, which, while noticeable, was considered acceptable as the primary objects remained recognizable.

\textit{Winter} modifications presented issues such as unrealistic over-tinting and missing parts of bodies or objects, which reduced the visual quality of certain examples. In contrast, lighting modifications—such as \textit{sunny}, \textit{night}, and \textit{dawn} scenes—did not present significant issues, and all generated data passed the filtering process without major problems.

In summary, although CosXLEdit provides highly realistic and coherent transformations with a high acceptance rate, it is not free from failure cases. Despite these occasional errors, the model remains reliable for generating training data, even without human filtering. However, post-processing may be necessary for specific use cases to ensure the highest quality.

\subsection{Robustness Evaluations}
\label{subsec:robustness-evaluation}
We benchmarked both CNN-based and Transformer-based segmentation models pretrained on the Cityscapes dataset to assess their robustness against synthetic scene modifications. These modifications simulate real-world environmental conditions, including variations in weather, seasons, and lighting. The evaluation compares the models' performance under these conditions, highlighting their ability to maintain accuracy in challenging scenarios.

\subsubsection{Model Performance on Adverse Conditions}
Table \ref{tab:model_performance_miou_error} presents the mIoU performance of several models across various adverse conditions. The performance drop is calculated as the difference in mIoU between the original Cityscapes dataset and the modified datasets, offering insights into the general robustness of each model.

CNN-based models, such as DeepLabV3+, ICNet, and DDRNet, show noticeable performance degradation under adverse conditions. DeepLabV3+ experiences the largest performance decline, dropping from 79.6 to an average mIoU of 51.2. ICNet, which has the lowest original mIoU of 76.3, performs relatively well in adverse conditions, with an average mIoU of 61.0, while DDRNet falls to an average mIoU of 59.3. Interestingly, despite its lower baseline, ICNet shows smaller drops in performance in several adverse datasets compared to other CNN models.

Transformer-based models demonstrate greater resilience. SegFormer, starting with a baseline mIoU of 76.5, achieves an average mIoU of 61.2 across conditions. SETR maintains an average mIoU of 68.0 across all adverse conditions, with a modest performance drop of 9.8 points. Mask2Former, with the highest baseline mIoU of 82.6, sustains strong performance in adverse conditions, with an average drop of 14.3 points, resulting in an average mIoU of 68.2.

Overall, CNN-based models exhibit an average performance drop of 21.4 points, while Transformer-based models show a smaller drop of 12.8 points. Transformers consistently maintain better performance and robustness across various environmental conditions, aligning with previous findings that Transformers handle diverse scenarios more effectively than CNNs \cite{vit_robustness}.

\begin{table*}
\centering
\caption{Model Performance (mIoU) with Performance Drops across Various Adverse Conditions. The table evaluates the robustness of models under different weather (rainy, foggy), seasonal (spring, autumn, snow), and lighting (sunny, night, dawn) conditions.}
\label{tab:model_performance_miou_error}
\resizebox{\textwidth}{!}{
\begin{tabular}{|l|c|cc|ccc|ccc|c|}
\toprule
\multirow{2}{*}{\textbf{Model Name}} & \multirow{2}{*}{\textbf{Ori.}} & \multicolumn{2}{c|}{\textbf{Weathers}} & \multicolumn{3}{c|}{\textbf{Seasons}} & \multicolumn{3}{c|}{\textbf{Lightings}} & \textbf{Adverse} \\ 
& & \textbf{Rainy} & \textbf{Foggy} & \textbf{Spring} & \textbf{Autumn} & \textbf{Snow} & \textbf{Sunny} & \textbf{Night} & \textbf{Dawn} & \textbf{Avg.} \\
\midrule
DeepLabV3+ & 79.6 & 58.2 (21.4) & 66.1 (13.5) & 69.5 (10.1) & 46.8 (32.9) & 22.3 (57.3) & 42.6 (37.0) & 47.4 (32.2) & 57.7 (21.9) & 51.2 (28.4) \\
ICNet & 76.3 & 68.8 (~7.5) & 69.3 (~7.0) & 70.6 (~5.7) & 66.7 (~9.6) & 27.0 (49.3) & 54.7 (21.6) & 61.2 (15.1) & 67.5 (~8.8) & 61.0 (15.2) \\ 
DDRNet & 80.0 & 65.6 (14.4) & 69.3 (10.7) & 71.9 (~8.1) & 55.1 (24.9) & 30.2 (49.8) & 53.5 (26.5) & 58.9 (21.1) & 64.8 (15.2) & 59.3 (20.6) \\
\cmidrule{1-11}
\multicolumn{1}{|c|}{CNN Average} & 78.6 & 64.2 (14.5) & 68.2 (10.4) & 70.7 (~7.9) & 56.2 (22.5) & 26.5 (52.1) & 50.2 (28.4) & 55.8 (22.8) & 63.3 (15.3) & 57.2 (21.4) \\
\midrule
SegFormer & 76.5 & 65.9 (10.7) & 65.7 (10.8) & 69.0 (7.5) & 65.4 (11.1) & 46.9 (29.6) & 57.9 (18.7) & 63.4 (13.2) & 66.1 (10.5) & 61.2 (16.0) \\
SETR & 77.6 & 71.8 (~5.8) & 70.6 (~6.9) & 73.2 (~4.4) & 71.3 (~6.3) & 61.4 (16.2) & 61.9 (15.7) & 70.3 (~7.3) & 73.0 (~4.6) & 68.0 (~9.8) \\ 
Mask2Former & 82.6 & 67.7 (14.9) & 72.9 (~9.7) & 74.9 (~7.6) & 73.7 (~8.8) & 62.8 (19.8) & 63.0 (19.5) & 69.0 (13.6) & 75.0 (~7.6) & 68.2 (14.3) \\
\cmidrule{1-11}
\multicolumn{1}{|c|}{Trans. Average} & 78.9 & 68.5 (10.4) & 69.8 (~9.1) & 72.4 (~6.5) & 70.1 (~8.8) & 57.1 (21.8) & 60.9 (18.0) & 67.5 (11.4) & 71.3 (~7.6) & 66.1 (12.8) \\
\midrule
\multicolumn{1}{|c|}{All Average} & 78.8 & 66.3 (12.5) & 69.0 (~9.8) & 71.5 (~7.2) & 63.2 (15.6) & 41.8 (37.0) & 55.6 (23.2) & 61.7 (17.1) & 67.3 (11.4) & 61.2 (17.6) \\
\bottomrule
\end{tabular}
}
\end{table*}

\subsubsection{IoU Drop Across Adverse Conditions and Classes}

\Figure[]()[width=\linewidth]{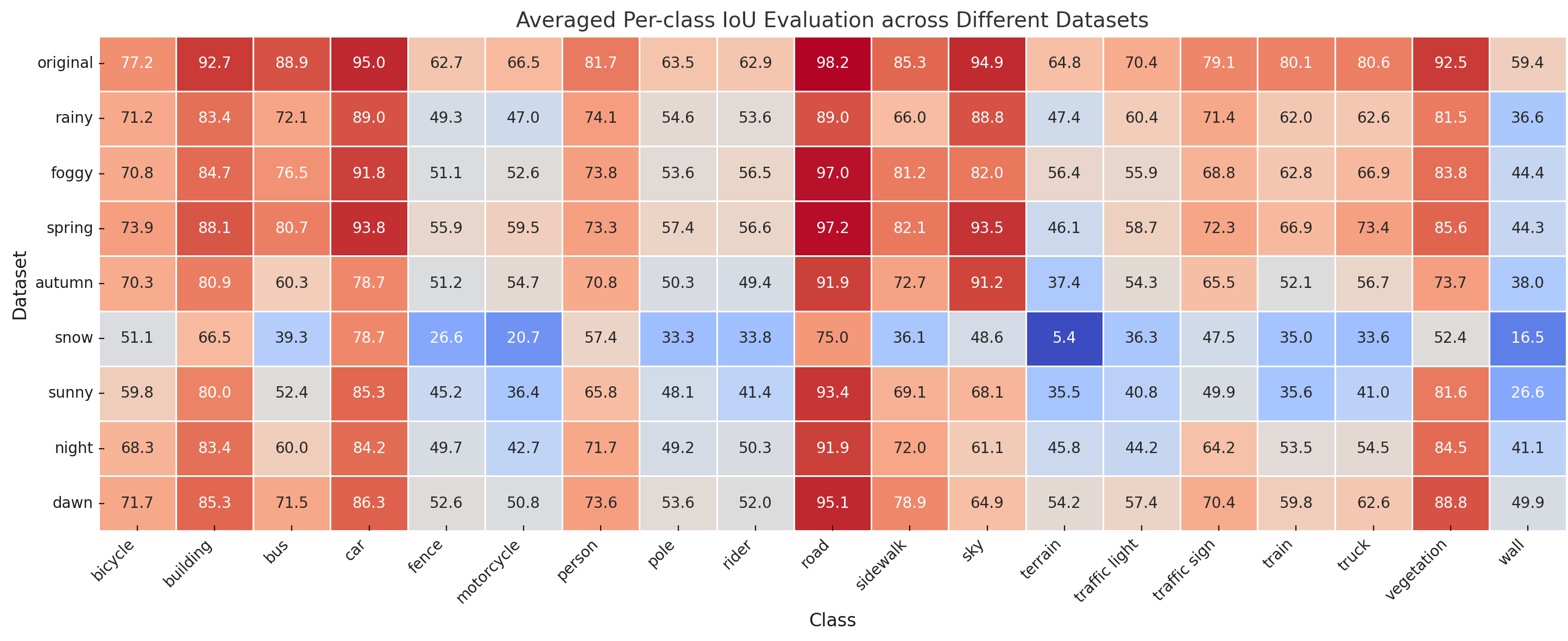}
{Per-class IoU evaluation across different adverse conditions. The heatmap illustrates significant IoU drops for classes like terrain and motorcycle in snow and sunny conditions. \label{fig:per_class_evaluation}}

\Figure[]()[width=\linewidth]{images/IoU_drop_distribution.png}
{Distribution of IoU drops for each dataset and class. The violin plots show snow and sunny conditions result in the largest IoU drops across datasets, while the bus, train, and motorcycle classes exhibit the largest drops across all conditions. In contrast, rainy, foggy, and spring conditions show more consistent performance across most classes. \label{fig:IoU_error}}

Figure \ref{fig:per_class_evaluation} presents the per-class IoU scores across different adverse conditions, providing insights into how various modifications impact object classes. \textit{Snow} and \textit{sunny} conditions lead to the largest performance drops, particularly for classes like terrain and motorcycle. For example, the terrain class drops from 64.8 mIoU in the original dataset to 5.4 mIoU in \textit{snow}, and the motorcycle class falls from 66.5 to 20.7. In contrast, more robust classes such as car, building, and road maintain relatively higher IoU scores across all conditions, with smaller performance drops.

\textit{Snow} proves to be the most challenging condition for many object classes, likely due to the drastic change in scene textures and visibility, making it harder for models to distinguish objects. On the other hand, \textit{rainy} and \textit{foggy} conditions result in more moderate declines in performance, suggesting that models are better equipped to handle these weather conditions than extreme environments like \textit{snow}.

Figure \ref{fig:IoU_error} shows the distribution of IoU drops for each dataset and class. The violin plots reveal that \textit{snow}, \textit{sunny}, \textit{night}, and \textit{autumn} conditions exhibit the largest IoU drops, especially for classes like train, terrain, and bus. Conversely, \textit{rainy}, \textit{foggy}, and \textit{spring} conditions exhibit narrower distributions of IoU drops, indicating more consistent model performance across most classes.

Overall, \textit{snow} and \textit{sunny} conditions present the most significant challenges for current segmentation models, with classes like terrain, bus, and motorcycle showing the largest performance drops. These findings suggest that improving model performance in extreme environments like snow should be a priority, especially in regions with diverse weather conditions. In contrast, models demonstrate greater robustness for classes such as car and road, which exhibit smaller IoU drops across all datasets.

\Figure[]()[width=\linewidth]{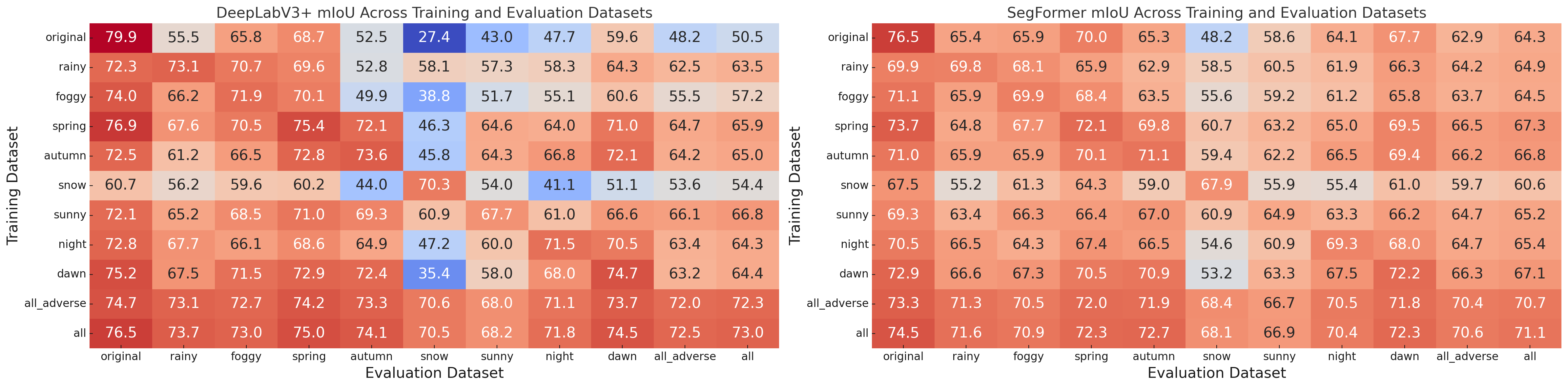}
{mIoU Performance of DeepLabV3+ and SegFormer across Various Training and Evaluation Datasets. The heatmap shows the mIoU performance of each model when trained on one dataset and evaluated across multiple environments, highlighting the generalization of the models to both adverse and non-adverse conditions. \label{fig:cross-dataset-evaluation}}

\subsection{Cross-dataset Evaluations}
\label{subsec:cross-dataset-evaluations}
We evaluate the performance of DeepLabV3+ and SegFormer across a range of environmental conditions using the Cityscape-Adverse dataset. These models were chosen due to their significant performance drops in prior evaluations under adverse conditions. We aim to assess how training these models on diverse adverse datasets impacts their robustness and generalization capabilities. All models were trained using their original hyperparameters and training steps.

\subsubsection{mIoU Performance Across Environmental Conditions}
\label{subsubsec:model_adverse_performance}
We analyze how different training datasets influence evaluation performance across adverse environments, as shown in Figure \ref{fig:cross-dataset-evaluation}. Specifically, we examine the impact of training on synthetic adverse datasets, including comprehensive sets like \textit{all\_adverse} and \textit{all}, to understand their effect on model generalization.

While comprehensive training improves generalization, models trained on the \textit{original} dataset still perform best when evaluated on it. For example, DeepLabV3+ achieves an mIoU of 79.9 on \textit{original}, and SegFormer achieves 76.5. However, both models experience significant performance drops in adverse environments, highlighting the trade-off between domain-specific optimization and broader generalization.

Training on individual adverse datasets, such as \textit{rainy} or \textit{foggy}, improves generalization within similar conditions. DeepLabV3+ trained on \textit{rainy} achieves 70.67 mIoU on \textit{foggy}, while SegFormer achieves 68.1. Although training on adverse datasets improves adaptability within related environments, performance on the \textit{original} dataset remains lower compared to models trained exclusively on it.

Comprehensive training with datasets like \textit{all\_adverse} or \textit{all} improves generalization across a broader range of conditions. For example, DeepLabV3+ trained on \textit{all} achieves 70.5 mIoU on \textit{snow}, and SegFormer trained on \textit{all\_adverse} achieves 70.51 on \textit{night}. However, models trained on comprehensive datasets often perform worse on the \textit{original} dataset, likely because training is spread across a wider range of conditions, diluting the model’s focus on specific environments.

This trade-off suggests potential improvements. Strategies such as increasing model capacity, extending training steps, or employing continual learning \cite{umberto_michieli__2019, arthur_douillard__2021} could enhance performance across diverse conditions. Although these strategies are beyond the scope of this paper, they warrant further exploration in future research.

In summary, comprehensive training on datasets like \textit{all\_adverse} or \textit{all} enhances generalization across both adverse and non-adverse environments. However, this comes at the cost of reduced performance on the \textit{original} dataset, underscoring the trade-off between specificity and robustness. Future work could explore strategies to mitigate this trade-off and improve model performance, especially in challenging conditions like \textit{snow} and \textit{sunny}.

\Figure[]()[width=\linewidth]{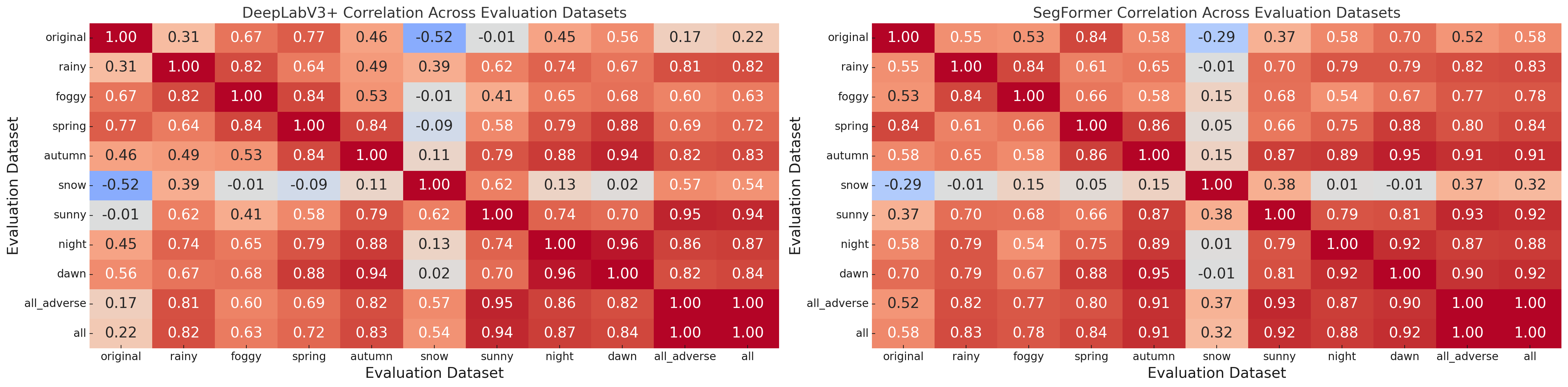}
{Correlation of mIoU Values across Evaluation Datasets for DeepLabV3+ and SegFormer. The heatmap illustrates the correlation of model performance across different environmental conditions, revealing generalization patterns and potential performance gaps in adverse environments. \label{fig:cross-dataset-correlation}}

\subsubsection{Correlation of mIoU Across Dataset}
\label{subsubsec:correlation_analysis}

We examine the correlation of mIoU values across different evaluation datasets to assess how well models generalize under various conditions. The correlation matrices for DeepLabV3+ and SegFormer, shown in Figure \ref{fig:cross-dataset-correlation}, illustrate how performance transitions between adverse environments and the \textit{original} dataset.

For DeepLabV3+, strong correlations are observed between similar adverse datasets, such as \textit{spring} and \textit{dawn}, indicating that the model generalizes well across visually consistent environments. However, performance drops significantly when transitioning from the \textit{original} dataset to challenging conditions like \textit{snow}, where the correlation is negative. Low correlations between the \textit{original} dataset and \textit{night} further indicate limited generalization to visually distinct environments.

SegFormer shows a similar pattern, with generally higher correlations across adverse conditions, indicating stronger adaptability. For instance, correlations between datasets like \textit{spring} and \textit{autumn} highlight consistent performance in related environments. SegFormer also demonstrates better generalization across adverse environments, such as between \textit{rainy} and \textit{night}, suggesting resilience in handling diverse conditions.

Both models struggle with extreme conditions like \textit{snow}. DeepLabV3+ shows a negative correlation between the \textit{original} dataset and \textit{snow}, while SegFormer shows similarly low correlations between \textit{snow} and other environments, reflecting the difficulty of transitioning between visually distinct domains.

Comprehensive training improves generalization for both models. Correlations between \textit{all\_adverse} and \textit{sunny} are high for both DeepLabV3+ and SegFormer, demonstrating that exposure to diverse adverse conditions enhances performance across a broad range of datasets. However, challenges remain in adapting to extreme environments like \textit{snow}.

In conclusion, both models generalize well across similar adverse conditions, with SegFormer, a transformer-based model, demonstrating stronger adaptability. However, transitions between highly distinct environments, such as from the \textit{original} dataset to \textit{snow}, remain challenging. Comprehensive training improves generalization, but achieving balanced performance across diverse environments requires further investigation.

\Figure[]()[width=\linewidth]{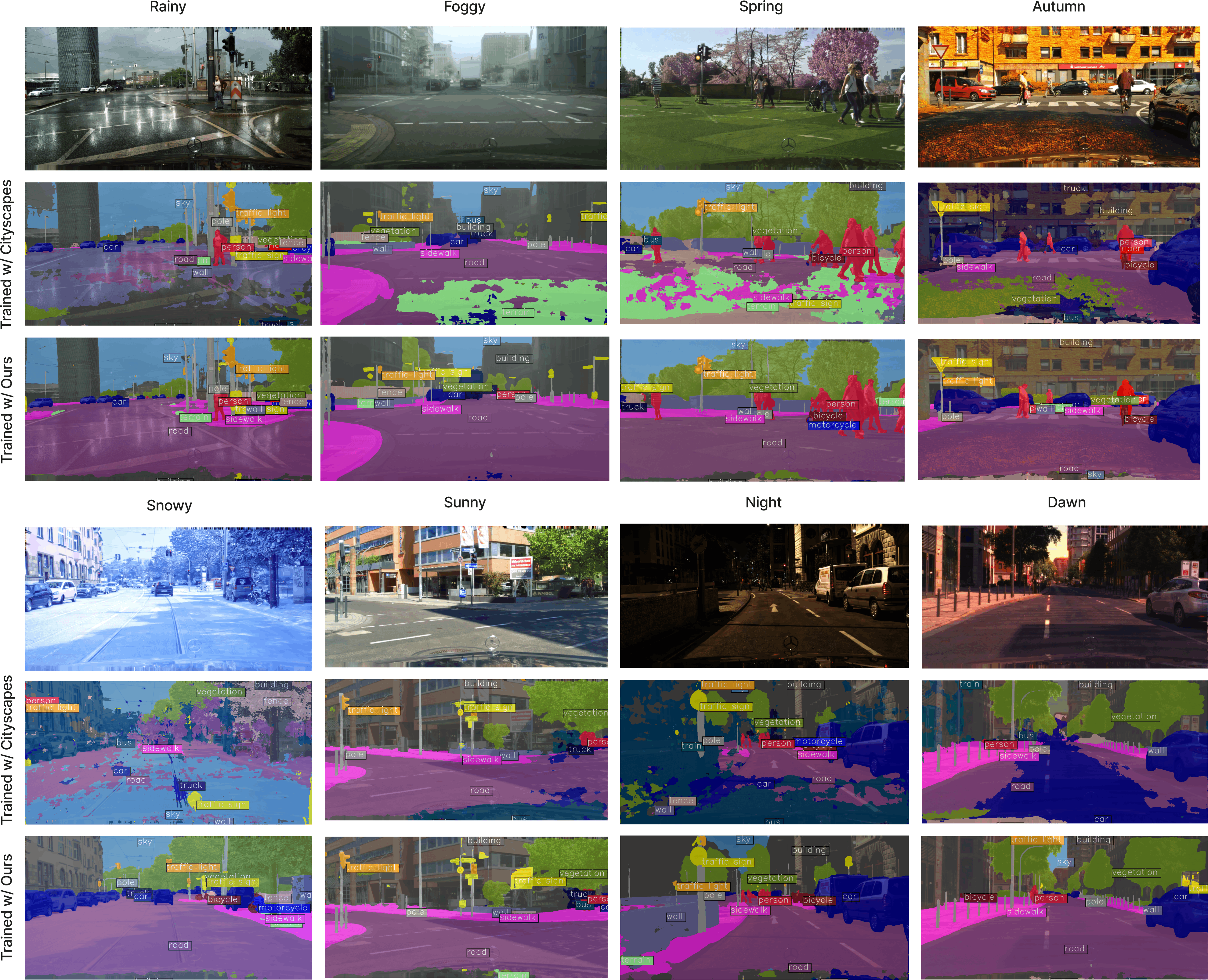}
{Segmentation results when the DeepLabV3+ model is trained with our dataset compared to the original Cityscapes dataset.
\label{fig:training-dataset-comparison}}

\subsubsection{Qualitative Comparison of Original vs Cityscape-Adverse}

The results in Figure \ref{fig:training-dataset-comparison} demonstrate significant improvements in segmentation performance across various environmental conditions when the DeepLabV3+ model is trained with our Cityscape-Adverse dataset. Specifically, training with our dataset enhances the model's ability to accurately delineate object boundaries and classify objects in challenging scenarios, such as \textit{foggy}, \textit{night}, \textit{rainy}, and \textit{snow} conditions. For example, in the \textit{night} and \textit{snow} cases, our dataset enables more precise segmentation, distinguishing between road elements, vehicles, and pedestrians, where the model trained on the original Cityscapes dataset struggles.

Our dataset is particularly effective in complex scenes. In the \textit{spring} condition, the model adeptly handles multiple objects, such as pedestrians and bicycles, without the overlap issues present in the model trained on the original dataset. However, under ideal conditions like \textit{sunny}, the improvement is less pronounced, suggesting that the original Cityscapes dataset is already sufficient for handling clear, well-lit scenes.

Nevertheless, the Cityscape-Adverse dataset provides substantial gains in segmentation quality, particularly in adverse or complex environments where the original dataset falls short. These results demonstrate that training with our dataset improves the model's robustness and adaptability, enabling it to perform better under diverse and challenging conditions.

\subsection{Robustness Enhancements with Synthetic Data}
\label{subsec:robustness-with-synthetic-data}

We investigate how synthetic data generated through diffusion-based image editing enhances the robustness and generalization of semantic segmentation models under real-world adverse conditions. Our evaluation focuses on datasets featuring a variety of challenging environments, particularly ACDC \cite{sakaridis_acdc_2021}, which includes subsets for \textit{snow}, \textit{rain}, \textit{fog}, and \textit{night}. This breakdown allows us to isolate the impact of specific adverse conditions on model performance.

Additionally, we evaluate models on a range of real-world datasets, including BDD100K \cite{yu2020bdd100k}, Dark Zurich \cite{SDV20}, Night Driving \cite{dai2018dark}, ApolloScape \cite{Huang_2020}, and Mapillary Vistas v2.0 \cite{vistasv2}, providing a broader spectrum of scenarios to assess model generalization.

To test the effectiveness of our synthetic dataset, we trained models in five configurations: (i) the original Cityscapes dataset (CS) \cite{cordts2016cityscapes}, (ii) Cityscapes augmented with GenVal data (CSGenVal) \cite{loiseau2024reliability}, (iii) a combination of Cityscapes and CSGenVal (CS+CSGenVal), (iv) our Cityscapes-Adverse dataset (CSAdv), and (v) a combination of Cityscapes and CSAdv (CS+CSAdv). This setup evaluates the models’ ability to generalize to unseen adverse conditions without relying on real-world adverse training data.

\subsubsection{Evaluation on ACDC Dataset}

The ACDC dataset serves as a primary benchmark for evaluating model robustness. Table \ref{tab:acdc-results} compares the performance of CNN-based and Transformer-based models trained on different configurations and evaluated on ACDC \cite{sakaridis_acdc_2021}.

CNN-based models show significant improvements when trained on CSAdv. For example, DeepLabV3+ trained on CSAdv achieves an average mIoU of 48.9 across adverse conditions, a 6.3-point increase over the baseline CS model’s 42.6 mIoU. Similarly, ICNet and DDRNet trained on CSAdv improve by 5.8 and 7.1 points, respectively. Notably, the most substantial gains occur under \textit{night} and \textit{snow} conditions. DDRNet trained on CS+CSAdv improves its mIoU in \textit{night} conditions by 23.5 points compared to the baseline.

In contrast, models trained on CSGenVal or CS+CSGenVal often underperform compared to the baseline CS model. DeepLabV3+ trained on CSGenVal shows a 12.3-point drop in mIoU, from 42.6 to 30.3. This suggests that GenVal’s synthetic data lacks the level of realism needed for robustness in complex adverse conditions, leading to reduced performance in challenging environments.

Transformer-based models also benefit from training on CSAdv. SegFormer improves its average mIoU by 4.7 points, from 47.4 to 52.1. Mask2Former and SETR show similar gains, with mIoU increases of 4.3 and 4.1 points, respectively. These models, already strong on the baseline dataset, demonstrate that training with CSAdv improves robustness without sacrificing performance on \textit{clean} conditions.

Overall, the CSAdv dataset provides significant performance improvements in adverse conditions for both CNN and Transformer models. Average mIoU improvements of 6.4 points for CNNs and 4.3 points for Transformers underscore its effectiveness, particularly in \textit{night} and \textit{snow}.

\begin{table*}[]
\centering
\caption{Performance Comparison of CNN and Transformer-based Models Trained on Different Configurations (CS: Cityscapes \cite{cordts2016cityscapes}, CSGenVal: Cityscape-GenVal \cite{loiseau2024reliability}, CS+CSGenVal: Cityscape with Cityscape-GenVal \cite{loiseau2024reliability}, CSAdv: Cityscapes-Adverse (Ours), and CS+CSAdv: Cityscape with Cityscape-Adverse (Ours)). Evaluated on ACDC \cite{sakaridis_acdc_2021} Datasets. The table presents the mIoU scores and performance enhancement (Enh.) for each model across clean, snow, rain, fog, and night conditions, along with the average mIoU across all adverse conditions (Adv. Avg.).}

\label{tab:acdc-results}
\begin{tabular}{|l|l|cc|cc|cc|cc|cc|cc|}
\toprule
\multirow{2}{*}{\textbf{Model Name}} & \multirow{2}{*}{\textbf{Trained On}} & \multicolumn{2}{c|}{\textbf{Clean}} & \multicolumn{2}{c|}{\textbf{Snow}} & \multicolumn{2}{c|}{\textbf{Rain}} & \multicolumn{2}{c|}{\textbf{Fog}} & \multicolumn{2}{c|}{\textbf{Night}} & \multicolumn{2}{c|}{\textbf{Adv. Avg}} \\ 
 &  & \textbf{mIoU} & \textbf{Enh.} & \textbf{mIoU} & \textbf{Enh.} & \textbf{mIoU} & \textbf{Enh.} & \textbf{mIoU} & \textbf{Enh.} & \textbf{mIoU} & \textbf{Enh.} & \textbf{mIoU} & \textbf{Enh.} \\ \midrule

\multirow{5}{*}{DeepLabV3+} 
    & CS \cite{cordts2016cityscapes}              & 58.8  & 0.0  & 44.3  & 0.0  & 50.3  & 0.0  & 63.7  & 0.0  & 12.2  & 0.0  & 42.6  & 0.0 \\ \cmidrule{2-14}
    & CSGenVal \cite{loiseau2024reliability}          & 39.9  & -18.9  & 26.8  & -17.5  & 33.7  & -16.6  & 45.8  & -17.9  & 15.0  & 2.8  & 30.3  & -12.3 \\  
    & CS+CSGenVal \cite{loiseau2024reliability}       & 47.4  & -11.4  & 35.8  & -8.5  & 41.7  & -8.6  & 51.9  & -11.8  & 20.1  & 7.9  & 37.4  & -5.3 \\ \cmidrule{2-14}
    & CSAdv (Ours)           & 62.4  & 3.6  & 49.3  & 5.0  & \textbf{50.7}  & \textbf{0.4}  & \textbf{64.9}  & \textbf{1.2}  & \textbf{30.7}  & \textbf{18.5}  & \textbf{48.9}  & \textbf{6.3} \\  
    & CS+CSAdv (Ours)        & \textbf{63.5}  & \textbf{4.7}  & \textbf{50.8}  & \textbf{6.5}  & 49.7  & -0.6  & 62.7  & -1.0  & 30.3  & 18.1  & 48.4  & 5.8 \\ \midrule

\multirow{5}{*}{ICNet} 
    & CS \cite{cordts2016cityscapes}              & \textbf{61.2}  & \textbf{0.0}  & 45.4  & 0.0  & 49.2  & 0.0  & 64.8  & 0.0  & 24.6  & 0.0  & 46.0  & 0.0 \\ \cmidrule{2-14}
    & CSGenVal \cite{loiseau2024reliability}          & 38.8  & -22.4  & 28.3  & -17.1  & 33.4  & -15.8  & 48.0  & -16.8  & 16.7  & -7.9  & 31.6  & -14.4 \\  
    & CS+CSGenVal \cite{loiseau2024reliability}       & 49.7  & -11.5  & 34.8  & -10.6  & 41.8  & -7.4  & 51.3  & -13.5  & 22.5  & -2.1  & 37.6  & -8.4 \\ \cmidrule{2-14}
    & CSAdv (Ours)           & \textbf{61.2}  & \textbf{0.0}  & \textbf{53.4}  & \textbf{8.0}  & 50.3  & 1.1  & 65.1  & \textbf{0.3}  & \textbf{38.3}  & \textbf{13.7}  & \textbf{51.8}  & \textbf{5.8} \\  
    & CS+CSAdv (Ours)        & 59.8  & -1.4  & \textbf{53.4}  & \textbf{8.0}  & \textbf{50.6}  & \textbf{1.4}  & \textbf{66.1}  & \textbf{1.3}  & 35.9  & 11.3  & 51.5  & 5.5 \\ \midrule

\multirow{5}{*}{DDRNet} 
    & CS \cite{cordts2016cityscapes}              & 58.4  & 0.0  & 42.5  & 0.0  & 48.0  & 0.0  & 64.5  & 0.0  & 10.6  & 0.0  & 41.4  & 0.0 \\ \cmidrule{2-14}
    & CSGenVal \cite{loiseau2024reliability}          & 34.1  & -24.3  & 24.0  & -18.5  & 30.3  & -17.7  & 41.5  & -23.0  & 16.8  & 6.2  & 28.2  & -13.3 \\  
    & CS+CSGenVal \cite{loiseau2024reliability}       & 48.4  & -10.0  & 36.2  & -6.3  & 40.5  & -7.5  & 52.2  & -12.3  & 21.7  & 11.1  & 37.7  & -3.8 \\ \cmidrule{2-14}
    & CSAdv (Ours)           & 59.1  & 0.7  & 44.3  & 1.8  & 49.0  & 1.0  & \textbf{65.8}  & \textbf{1.3}  & \textbf{34.8}  & \textbf{24.2}  & 48.5  & 7.1 \\  
    & CS+CSAdv (Ours)        & \textbf{61.2}  & \textbf{2.8}  & \textbf{47.5}  & \textbf{5.0}  & \textbf{51.3}  & \textbf{3.3}  & 65.1  & 0.6  & 34.1  & 23.5  & \textbf{49.5}  & \textbf{8.1} \\ \midrule \midrule

\multirow{5}{*}{CNN Average} 
        & CS \cite{cordts2016cityscapes}              & 59.5  & 0.0  & 44.1  & 0.0  & 49.2  & 0.0  & 64.3  & 0.0  & 15.8  & 0.0  & 43.3  & 0.0 \\ \cmidrule{2-14}
    & CSGenVal \cite{loiseau2024reliability}          & 37.6  & -21.9  & 26.4  & -17.7  & 32.5  & -16.7  & 45.1  & -19.2  & 16.2  & 0.4  & 30.0  & -13.3 \\  
    & CS+CSGenVal \cite{loiseau2024reliability}       & 48.5  & -11.0  & 35.6  & -8.5  & 41.3  & -7.8  & 51.8  & -12.5  & 21.4  & 5.6  & 37.5  & -5.8 \\ \cmidrule{2-14}
    & CSAdv (Ours)           & 60.9  & 1.4  & 49.0  & 4.9  & 50.0  & 0.8  & 65.3  & 0.9  & 34.6  & 18.8  & 49.7  & 6.4 \\  
    & CS+CSAdv (Ours)        & \textbf{61.5}  & \textbf{2.0}  & \textbf{50.6}  & \textbf{6.5}  & \textbf{50.5}  & \textbf{1.4}  & 64.6  & 0.3  & 33.4  & 17.6  & \textbf{49.8}  & \textbf{6.5} \\ \midrule \midrule

\multirow{5}{*}{SegFormer} 
    & CS \cite{cordts2016cityscapes}              & 57.2  & 0.0  & 51.2  & 0.0  & \textbf{52.7}  & \textbf{0.0}  & 67.1  & 0.0  & 18.5  & 0.0  & 47.4  & 0.0 \\ \cmidrule{2-14}
    & CSGenVal \cite{loiseau2024reliability}          & 38.6  & -18.6  & 29.6  & -21.6  & 35.1  & -17.6  & 47.7  & -19.4  & 17.4  & -1.1  & 32.5  & -14.9 \\  
    & CS+CSGenVal \cite{loiseau2024reliability}       & 46.9  & -10.3  & 35.6  & -15.6  & 42.0  & -10.7  & 51.0  & -16.1  & 22.1  & 3.6  & 37.7  & -9.7 \\ \cmidrule{2-14}
    & CSAdv (Ours)           & \textbf{60.9}  & \textbf{3.7}  & \textbf{54.7}  & \textbf{3.5}  & 49.6  & -3.1  & \textbf{69.3}  & \textbf{2.2}  & \textbf{34.6}  & \textbf{16.1}  & \textbf{52.1}  & \textbf{4.7} \\  
    & CS+CSAdv (Ours)        & 60.7  & 3.5  & 53.5  & 2.3  & 49.8  & -2.9  & \textbf{69.3}  & \textbf{2.2}  & 32.1  & 13.6  & 51.2  & 3.8 \\ \midrule

\multirow{5}{*}{Mask2Former} 
    & CS \cite{cordts2016cityscapes}              & \textbf{67.5}  & 0.0  & 55.4  & 0.0  & 58.2  & 0.0  & \textbf{73.8}  & 0.0  & 31.4  & 0.0  & 54.7  & 0.0 \\ \cmidrule{2-14}
    & CSGenVal \cite{loiseau2024reliability}          & 39.3  & -28.2  & 32.8  & -22.6  & 38.4  & -19.8  & 52.0  & -21.8  & 24.9  & -6.5  & 37.0  & -17.7 \\  
    & CS+CSGenVal \cite{loiseau2024reliability}       & 48.4  & -19.1  & 39.3  & -16.1  & 45.1  & -13.1  & 53.9  & -19.9  & 28.3  & -3.1  & 41.7  & -13.1 \\ \cmidrule{2-14}
    & CSAdv (Ours)           & \textbf{69.8}  & \textbf{2.3}  & 61.6  & 6.2  & \textbf{58.6}  & \textbf{0.4}  & 72.9  & -0.9  & 42.8  & 11.4  & \textbf{59.0}  & \textbf{4.3} \\  
    & CS+CSAdv (Ours)        & 68.7  & 1.2  & \textbf{62.4}  & \textbf{7.0}  & 55.3  & -2.9  & 73.0  & -0.8  & \textbf{43.7}  & \textbf{12.3}  & 58.6  & 3.9 \\ \midrule

\multirow{5}{*}{SETR} 
    & CS \cite{cordts2016cityscapes}              & 60.7  & 0.0  & 55.6  & 0.0  & 58.3  & 0.0  & 70.1  & 0.0  & 38.1  & 0.0  & 55.5  & 0.0 \\ \cmidrule{2-14}
    & CSGenVal \cite{loiseau2024reliability}          & 40.9  & -19.8  & 34.1  & -21.5  & 37.8  & -20.5  & 48.7  & -21.4  & 26.2  & -11.9  & 36.7  & -18.8 \\  
    & CS+CSGenVal \cite{loiseau2024reliability}       & 49.1  & -11.6  & 40.6  & -15.0  & 46.2  & -12.1  & 52.7  & -17.4  & 30.5  & -7.6  & 42.5  & -13.0 \\ \cmidrule{2-14}
    & CSAdv (Ours)           & \textbf{65.5}  & \textbf{4.8}  & 61.7  & 6.1  & 58.0  & -0.3  & \textbf{74.9}  & \textbf{4.8}  & \textbf{43.9}  & \textbf{5.8}  & \textbf{59.6}  & \textbf{4.1} \\  
    & CS+CSAdv (Ours)        & 64.6  & 3.9  & \textbf{62.2}  & \textbf{6.6}  & \textbf{59.6}  & \textbf{1.3}  & 73.9  & 3.8  & 40.9  & 2.8  & 59.2  & 3.6 \\ \midrule \midrule

\multirow{5}{*}{Trans. Average} 
    & CS \cite{cordts2016cityscapes}              & 61.8  & 0.0  & 54.1  & 0.0  & \textbf{56.4}  & \textbf{0.0}  & 70.3  & 0.0  & 29.3  & 0.0  & 52.5  & 0.0 \\ \cmidrule{2-14}
    & CSGenVal \cite{loiseau2024reliability}          & 39.6  & -22.2  & 32.2  & -21.9  & 37.1  & -19.3  & 49.5  & -20.9  & 22.8  & -6.5  & 35.4  & -17.1 \\  
    & CS+CSGenVal \cite{loiseau2024reliability}       & 48.1  & -13.7  & 38.5  & -15.6  & 44.4  & -12.0  & 52.5  & -17.8  & 27.0  & -2.4  & 40.6  & -11.9 \\ \cmidrule{2-14}
    & CSAdv (Ours)           & 65.4  & 3.6  & 59.3  & 5.3  & 55.4  & -1.0  & 72.4  & 2.0  & 40.4  & 11.1  & 56.9  & 4.3 \\  
    & CS+CSAdv (Ours)        & \textbf{64.7}  & \textbf{2.9}  & \textbf{59.4}  & \textbf{5.3}  & 54.9  & -1.5  & \textbf{72.1}  & \textbf{1.7}  & \textbf{38.9}  & \textbf{9.6}  & \textbf{56.3}  & \textbf{3.8} \\ \midrule \bottomrule

\end{tabular}
\end{table*}

\subsubsection{Evaluation on Other Real-World Datasets}

Table \ref{tab:real_world_datasets} extends the evaluation to additional real-world adverse datasets for assessing model generalization across a variety of conditions.

For CNN-based models, training on CSAdv leads to substantial gains. DeepLabV3+, for example, shows an average mIoU improvement of 13.7 points, increasing from 26.6 to 40.3. This improvement is particularly pronounced on \textit{Dark Zurich} and \textit{Night Driving}, with mIoU gains of 19.8 and 30.6 points, respectively. ICNet and DDRNet also exhibit notable gains, with mIoU increases of 8.4 and 15.6 points.

Training with CS+CSAdv yields comparable improvements. For instance, DDRNet trained on CS+CSAdv achieves an average mIoU of 41.4, a 15.8-point increase over the baseline.

Transformer-based models similarly benefit from training on CSAdv. SegFormer improves its average mIoU by 6.3 points, rising from 34.4 to 40.7. Mask2Former and SETR also show gains, with mIoU increases of 4.4 and 2.8 points, respectively. These gains are consistent across most datasets, with notable performance improvements on \textit{Dark Zurich} and \textit{Night Driving}.

In contrast, models trained on CSGenVal or CS+CSGenVal generally underperform compared to the baseline CS model. For example, Mask2Former trained on CSGenVal shows a 10-point drop in mIoU, from 45.1 to 35.1, suggesting CSGenVal lacks the realism needed for complex real-world conditions, leading to underperformance.

In summary, our diffusion-based synthetic dataset significantly enhances the robustness and generalization of semantic segmentation models across diverse real-world adverse conditions. These improvements are particularly impactful in night-time and low-visibility environments, highlighting the value of our approach in addressing challenges posed by adverse scenarios.

\begin{table*}[]
\centering
\caption{Performance Comparison of Various Models Trained on Different Configurations (CS: Cityscapes \cite{cordts2016cityscapes}, CSGenVal: Cityscapes-GenVal \cite{loiseau2024reliability}, CS+CSGenVal: Cityscapes with Cityscapes-GenVal, CSAdv: Cityscapes-Adverse (Ours), and CS+CSAdv: Cityscapes with Cityscapes-Adverse (Ours)). Evaluated on multiple datasets. The table presents the mIoU scores and performance enhancement (Enh.) for each model across BDD100K, Dark Zurich, Night Driving, ApolloScape, Mapillary V2, and the average mIoU across all datasets.}

\label{tab:real_world_datasets}
\begin{tabular}{|l|l|cc|cc|cc|cc|cc|cc|}
\toprule
\multirow{2}{*}{\textbf{Model Name}} & \multirow{2}{*}{\textbf{Trained On}} & \multicolumn{2}{c|}{\textbf{BDD100K}} & \multicolumn{2}{c|}{\textbf{Dark Zurich}} & \multicolumn{2}{c|}{\textbf{Night Driving}} & \multicolumn{2}{c|}{\textbf{ApolloScape}} & \multicolumn{2}{c|}{\textbf{Mapillary V2}} & \multicolumn{2}{c|}{\textbf{Average}} \\

 &  & \textbf{mIoU} & \textbf{Enh.} & \textbf{mIoU} & \textbf{Enh.} & \textbf{mIoU} & \textbf{Enh.} & \textbf{mIoU} & \textbf{Enh.} & \textbf{mIoU} & \textbf{Enh.} & \textbf{mIoU} & \textbf{Enh.} \\ \midrule

\multirow{5}{*}{DeepLabV3+} 
& CS \cite{cordts2016cityscapes} & 37.3 & 0.0 & 10.0 & 0.0 & 20.9 & 0.0 & 37.3 & 0.0 & 38.3 & 0.0 & 26.6 & 0.0 \\ \cmidrule{2-14}
& CSGenVal \cite{loiseau2024reliability} & 31.0 & -6.3 & 16.6 & 6.6 & 30.0 & 9.0 & 29.8 & -7.5 & 28.7 & -9.6 & 26.3 & -0.4 \\
& CS+CSGenVal \cite{loiseau2024reliability} & 37.1 & -0.2 & 19.8 & 9.7 & 34.5 & 13.6 & 34.2 & -3.1 & 34.4 & -3.9 & 30.7 & 4.1 \\ \cmidrule{2-14}
& CSAdv (Ours) & \textbf{44.1} & \textbf{6.8} & \textbf{29.8} & \textbf{19.8} & \textbf{51.5} & \textbf{30.6} & \textbf{41.1} & \textbf{3.8} & 38.8 & 0.6 & \textbf{40.3} & \textbf{13.7} \\
& CS+CSAdv (Ours) & 43.5 & 6.2 & 29.8 & 19.7 & 48.6 & 27.7 & 41.7 & 4.4 & \textbf{39.9} & \textbf{1.6} & 40.0 & 13.3 \\ \midrule

\multirow{5}{*}{ICNet} 
& CS \cite{cordts2016cityscapes} & 44.1 & 0.0 & 24.3 & 0.0 & 42.2 & 0.0 & 32.1 & 0.0 & 37.1 & 0.0 & 33.9 & 0.0 \\ \cmidrule{2-14}
& CSGenVal \cite{loiseau2024reliability} & 30.5 & -13.6 & 19.7 & -4.6 & 25.9 & -16.3 & 23.4 & -8.7 & 23.1 & -14.1 & 23.0 & -10.9 \\
& CS+CSGenVal \cite{loiseau2024reliability} & 41.7 & -2.4 & 22.2 & -2.1 & 37.1 & -5.1 & 27.2 & -4.9 & 30.9 & -6.2 & 29.3 & -4.6 \\ \cmidrule{2-14}
& CSAdv (Ours) & 49.2 & 5.1 & \textbf{37.6} & \textbf{13.3} & \textbf{57.0} & \textbf{14.8} & 36.2 & 4.1 & 38.4 & 1.3 & \textbf{42.3} & \textbf{8.4} \\
& CS+CSAdv (Ours) & \textbf{50.4} & \textbf{6.3} & 36.3 & 12.0 & 55.0 & 12.8 & \textbf{35.9} & \textbf{3.8} & \textbf{38.7} & \textbf{1.6} & 41.5 & 7.5 \\ \midrule

\multirow{5}{*}{DDRNet} 
& CS \cite{cordts2016cityscapes} & 40.7 & 0.0 & 9.2 & 0.0 & 18.2 & 0.0 & 37.2 & 0.0 & 38.0 & 0.0 & 25.6 & 0.0 \\ \cmidrule{2-14}
& CSGenVal \cite{loiseau2024reliability} & 26.8 & -13.9 & 18.0 & 8.9 & 24.0 & 5.8 & 22.6 & -14.7 & 21.2 & -16.9 & 21.4 & -4.2 \\
& CS+CSGenVal \cite{loiseau2024reliability} & 36.8 & -3.9 & 21.3 & 12.2 & 33.7 & 15.5 & 33.0 & -4.3 & 32.8 & -5.2 & 30.2 & 4.6 \\ \cmidrule{2-14}
& CSAdv (Ours) & 46.7 & 6.0 & \textbf{34.3} & \textbf{25.1} & \textbf{52.5} & \textbf{34.4} & 39.6 & 2.4 & 38.5 & 0.5 & \textbf{41.2} & \textbf{15.6} \\
& CS+CSAdv (Ours) & \textbf{46.8} & \textbf{6.1} & 33.5 & 24.3 & 52.6 & 34.4 & \textbf{39.6} & \textbf{2.4} & \textbf{39.9} & \textbf{1.9} & 41.4 & 15.8 \\ \midrule \midrule

\multirow{5}{*}{CNN Average} 
& CS \cite{cordts2016cityscapes} & 40.7 & 0.0 & 14.5 & 0.0 & 27.1 & 0.0 & 35.5 & 0.0 & 37.8 & 0.0 & 28.7 & 0.0 \\ \cmidrule{2-14}
& CSGenVal \cite{loiseau2024reliability} & 29.4 & -11.3 & 18.1 & 3.6 & 26.6 & -0.5 & 25.3 & -10.3 & 24.3 & -13.5 & 23.6 & -5.2 \\
& CS+CSGenVal \cite{loiseau2024reliability} & 38.6 & -2.1 & 21.1 & 6.6 & 35.1 & 8.0 & 31.5 & -4.1 & 32.7 & -5.1 & 30.1 & 1.4 \\ \cmidrule{2-14}
& CSAdv (Ours) & \textbf{46.7} & \textbf{6.0} & \textbf{33.9} & \textbf{19.4} & \textbf{53.7} & \textbf{26.6} & \textbf{39.0} & \textbf{3.4} & 38.6 & 0.8 & \textbf{41.3} & \textbf{12.5} \\
& CS+CSAdv (Ours) & 46.9 & 6.2 & 33.2 & 18.7 & 52.1 & 25.0 & 39.1 & 3.5 & \textbf{39.5} & \textbf{1.7} & 40.9 & 12.2 \\ \midrule \midrule

\multirow{5}{*}{SegFormer} 
& CS \cite{cordts2016cityscapes} & 43.1 & 0.0 & 18.5 & 0.0 & 31.4 & 0.0 & 39.5 & 0.0 & 39.5 & 0.0 & 34.4 & 0.0 \\ \cmidrule{2-14}
& CSGenVal \cite{loiseau2024reliability} & 31.1 & -12.1 & 18.1 & -0.4 & 26.5 & -4.9 & 28.6 & -10.8 & 27.9 & -11.6 & 26.4 & -7.9 \\
& CS+CSGenVal \cite{loiseau2024reliability} & 34.9 & -8.2 & 21.3 & 2.7 & 33.0 & 1.6 & 32.4 & -7.1 & 35.7 & -3.9 & 31.5 & -2.9 \\ \cmidrule{2-14}
& CSAdv (Ours) & \textbf{44.6} & \textbf{1.5} & \textbf{32.7} & \textbf{14.2} & \textbf{48.1} & \textbf{16.7} & \textbf{37.7} & \textbf{-1.7} & \textbf{40.4} & \textbf{0.8} & \textbf{40.7} & \textbf{6.3} \\
& CS+CSAdv (Ours) & 45.0 & 1.9 & 31.7 & 13.2 & 48.6 & 17.1 & 38.5 & -0.9 & 40.3 & 0.7 & 40.8 & 6.4 \\ \midrule

\multirow{5}{*}{Mask2Former} 
& CS \cite{cordts2016cityscapes} & \textbf{52.6} & 0.0 & 27.2 & 0.0 & 53.3 & 0.0 & \textbf{46.5} & 0.0 & 45.9 & 0.0 & 45.1 & 0.0 \\ \cmidrule{2-14}
& CSGenVal \cite{loiseau2024reliability} & 39.5 & -13.0 & 26.6 & -0.6 & 40.7 & -12.6 & 35.0 & -11.5 & 33.7 & -12.2 & 35.1 & -10.0 \\
& CS+CSGenVal \cite{loiseau2024reliability} & 43.0 & -9.5 & 30.4 & 3.2 & 49.4 & -3.9 & 44.6 & -1.9 & 40.5 & -5.4 & 41.6 & -3.5 \\ \cmidrule{2-14}
& CSAdv (Ours) & 54.8 & 2.2 & 38.0 & 10.8 & \textbf{61.9} & \textbf{8.6} & 46.4 & -0.1 & 46.8 & 0.9 & \textbf{49.5} & \textbf{4.4} \\
& CS+CSAdv (Ours) & \textbf{55.3} & \textbf{2.8} & \textbf{38.5} & \textbf{11.3} & 65.5 & 12.2 & \textbf{47.2} & \textbf{0.7} & \textbf{46.5} & \textbf{0.6} & 50.6 & 5.5 \\ \midrule

\multirow{5}{*}{SETR} 
& CS \cite{cordts2016cityscapes} & \textbf{55.9} & 0.0 & 34.4 & 0.0 & 59.7 & 0.0 & 42.1 & 0.0 & 43.1 & 0.0 & 47.0 & 0.0 \\ \cmidrule{2-14}
& CSGenVal \cite{loiseau2024reliability} & 40.1 & -15.7 & 24.4 & -10.0 & 42.2 & -17.6 & 29.2 & -12.9 & 32.9 & -10.1 & 33.8 & -13.2 \\
& CS+CSGenVal \cite{loiseau2024reliability} & 45.7 & -10.2 & 26.3 & -8.1 & 44.1 & -15.6 & 33.5 & -8.6 & 37.9 & -5.2 & 37.5 & -9.5 \\ \cmidrule{2-14}
& CSAdv (Ours) & 56.0 & 0.2 & \textbf{41.6} & \textbf{7.2} & \textbf{61.2} & \textbf{1.5} & \textbf{42.0} & -0.1 & \textbf{44.4} & \textbf{1.3} & \textbf{49.8} & \textbf{2.8} \\
& CS+CSAdv (Ours) & 55.3 & -0.6 & 36.8 & 2.4 & 60.6 & 0.9 & 42.0 & -0.1 & 44.1 & 1.1 & 47.9 & 0.9 \\ \midrule \midrule

\multirow{5}{*}{Trans. Average} 
& CS \cite{cordts2016cityscapes} & \textbf{50.5} & 0.0 & 26.7 & 0.0 & 48.1 & 0.0 & 42.7 & 0.0 & 42.8 & 0.0 & 42.2 & 0.0 \\ \cmidrule{2-14}
& CSGenVal \cite{loiseau2024reliability} & 36.9 & -13.6 & 23.0 & -3.7 & 36.5 & -11.7 & 31.0 & -11.7 & 31.5 & -11.3 & 31.8 & -10.4 \\
& CS+CSGenVal \cite{loiseau2024reliability} & 41.2 & -9.3 & 26.0 & -0.7 & 42.2 & -6.0 & 36.8 & -5.9 & 38.0 & -4.8 & 36.8 & -5.4 \\ \cmidrule{2-14}
& CSAdv (Ours) & 51.8 & 1.3 & \textbf{37.4} & \textbf{10.7} & \textbf{57.0} & \textbf{8.9} & \textbf{42.1} & -0.6 & \textbf{43.8} & \textbf{1.0} & \textbf{46.5} & \textbf{4.3} \\
& CS+CSAdv (Ours) & \textbf{51.9} & \textbf{1.4} & 35.6 & 8.9 & 58.2 & 10.1 & 42.6 & -0.1 & 43.6 & 0.8 & 46.4 & 4.2 \\ \midrule \bottomrule

\end{tabular}
\end{table*}

\section{Discussion}
The introduction of Cityscape-Adverse as a benchmark for evaluating the robustness of semantic segmentation models under adverse conditions underscores the increasing need for testing models in realistic yet challenging environments. By leveraging diffusion-based image editing techniques, Cityscape-Adverse offers a scalable and efficient solution for generating highly realistic adverse scenarios without requiring extensive real-world data collection. Its ability to preserve semantic labels while introducing diverse environmental modifications makes it a practical and valuable tool for robustness testing, enabling comprehensive evaluations of model performance in conditions that closely resemble real-world scenarios, such as varying environment conditions.

\subsection{Key Findings}
Our experimental results provide several important insights:
\begin{itemize}
    \item \textbf{Performance under Adverse Conditions}: Across diverse weather, seasonal, and lighting variations, all tested models experienced significant performance degradation compared to their performance on the original Cityscapes dataset. CNN-based architectures struggled the most under extreme conditions like snow and night. In contrast, Transformer-based models, such as SegFormer and Mask2Former, demonstrated greater resilience, highlighting their superior ability to handle complex out-of-distribution (OOD) scenarios.

    \item \textbf{Cross-Dataset Generalization}: Models trained on subsets of Cityscape-Adverse datasets generalized well to similar environmental conditions, such as rainy and foggy or night and dawn. However, snow—representing a major domain shift—proved to be the most challenging condition for all models. Training on a combination of all adverse datasets was the most effective strategy for achieving broad generalization across diverse scenarios.

    \item \textbf{Effectiveness of Synthetic Data}: Training on realistic synthetic datasets generated through diffusion-based methods significantly improved model robustness in OOD scenarios. Models trained on Cityscape-Adverse showed notable performance gains on both synthetic benchmarks and real-world adverse datasets (e.g., ACDC, Dark Zurich), outperforming previous approaches. These results underscore the importance of synthetic data closely mimicking real-world conditions and demonstrate the potential of synthetic data augmentation for improving model generalization to unseen environments.
\end{itemize}

\subsection{Limitations}
Despite the promising results, our approach has several limitations:
\begin{itemize}
    \item \textbf{Synthetic Data Quality}: While diffusion-based image editing generates realistic scene modifications, some failure cases were observed, particularly under conditions like foggy, autumn, and snow. These modifications occasionally introduced visual artifacts or inconsistencies that compromised the realism and semantic consistency of the synthetic data. Future improvements could focus on reducing such artifacts to enhance overall data fidelity.

    \item \textbf{Potential Human Error in Filtering}: Although human filtering was applied to ensure the validity of the Cityscape-Adverse benchmark, the process is susceptible to error. Subjectivity in assessing the credibility of images and the single-stage filtering process may have allowed some false positives or negatives to pass. A multi-stage filtering process or a mechanism such as majority voting could mitigate this issue and improve consistency in the filtered dataset.

    \item \textbf{Focus on Environmental Conditions}: The benchmark primarily addresses environmental changes such as weather and lighting. This work did not consider other distribution shifts, including occlusions, motion blur, and sensor noise, which could limit its comprehensiveness in assessing robustness across broader challenges.

    \item \textbf{Trade-off Between Robustness and Ideal Performance}: While models trained on Cityscape-Adverse showed enhanced robustness in adverse conditions, they sometimes underperformed in ideal scenarios, indicating a trade-off that remains to be resolved. Exploring strategies to balance robustness in both adverse and ideal conditions could be a valuable area for future research.
\end{itemize}

\subsection{Future Work}
Several avenues for future work could extend the scope of our research:
\begin{itemize}
    \item \textbf{Extension to Additional OOD Scenarios}: Future research could focus on generating synthetic datasets that cover more complex OOD scenarios, such as dynamic environments with occlusions, motion blur, or sensor-specific noise. Including these conditions would provide a more comprehensive evaluation of model robustness across a wider variety of real-world challenges.

    \item \textbf{Advancements in Diffusion Models}: Our current approach relies on the zero-shot capabilities of a pre-trained diffusion model. Further advancements in diffusion models, including fine-tuning on real adverse datasets, could improve the quality and realism of the synthetic data. This could help mitigate failure cases, such as visual artifacts, and enhance the generalization performance of models trained on synthetic data.

    \item \textbf{Expanding Beyond Semantic Segmentation}: Extending Cityscape-Adverse to support tasks like object detection, instance segmentation, or depth estimation would provide broader insights into model robustness under adverse conditions across various computer vision tasks.

    \item \textbf{Generalization Across Diverse Semantic Shifts}: Future research could investigate generating synthetic data without constraining the original scene layout, facilitating generalization across datasets with differing semantic content (e.g., architectural differences between regions).

    \item \textbf{Synthetic and Real Data Integration}: Future research could explore advanced domain adaptation techniques and approaches to combine synthetic and real data more effectively. These techniques could help bridge the gap between real and synthetic environments, boosting performance across both domains.
\end{itemize}

\section{Conclusion}
In this paper, we introduced Cityscape-Adverse, a novel benchmark designed to evaluate the robustness of semantic segmentation models under a variety of adverse environmental conditions. By leveraging diffusion-based image editing techniques, we generated highly realistic synthetic datasets that simulate challenging real-world scenarios, such as different weather conditions and varying lighting environments, while preserving the semantic integrity of the original data.

Our experimental results demonstrated that all tested models, especially CNN-based architectures, experienced significant performance degradation in extreme conditions, such as snow and night. Transformer-based models, such as SegFormer and Mask2Former, exhibited greater resilience, highlighting their potential for handling out-of-distribution (OOD) scenarios more effectively. Furthermore, models trained on Cityscape-Adverse generalized well to real-world datasets like ACDC and Dark Zurich, underscoring the utility of synthetic data in improving model robustness.

While Cityscape-Adverse primarily focuses on environmental conditions, future work could expand the benchmark to address additional distribution shifts, such as occlusions, motion blur, and architectural differences. We also identified a trade-off between robustness in adverse conditions and performance in ideal scenarios, which presents an interesting challenge for further exploration.

Overall, Cityscape-Adverse provides a practical and scalable tool for robustness testing, bridging the gap between idealized datasets and real-world challenges. We believe this benchmark will facilitate the development of more resilient computer vision models, enabling better generalization across a wide range of environments and conditions. Future work will focus on extending the benchmark to other tasks, such as object detection and depth estimation, and exploring more advanced synthetic data generation techniques.


\bibliographystyle{IEEEtran}
\bibliography{ref}


\begin{IEEEbiography}[{\includegraphics[width=1in,height=1.25in,clip,keepaspectratio]{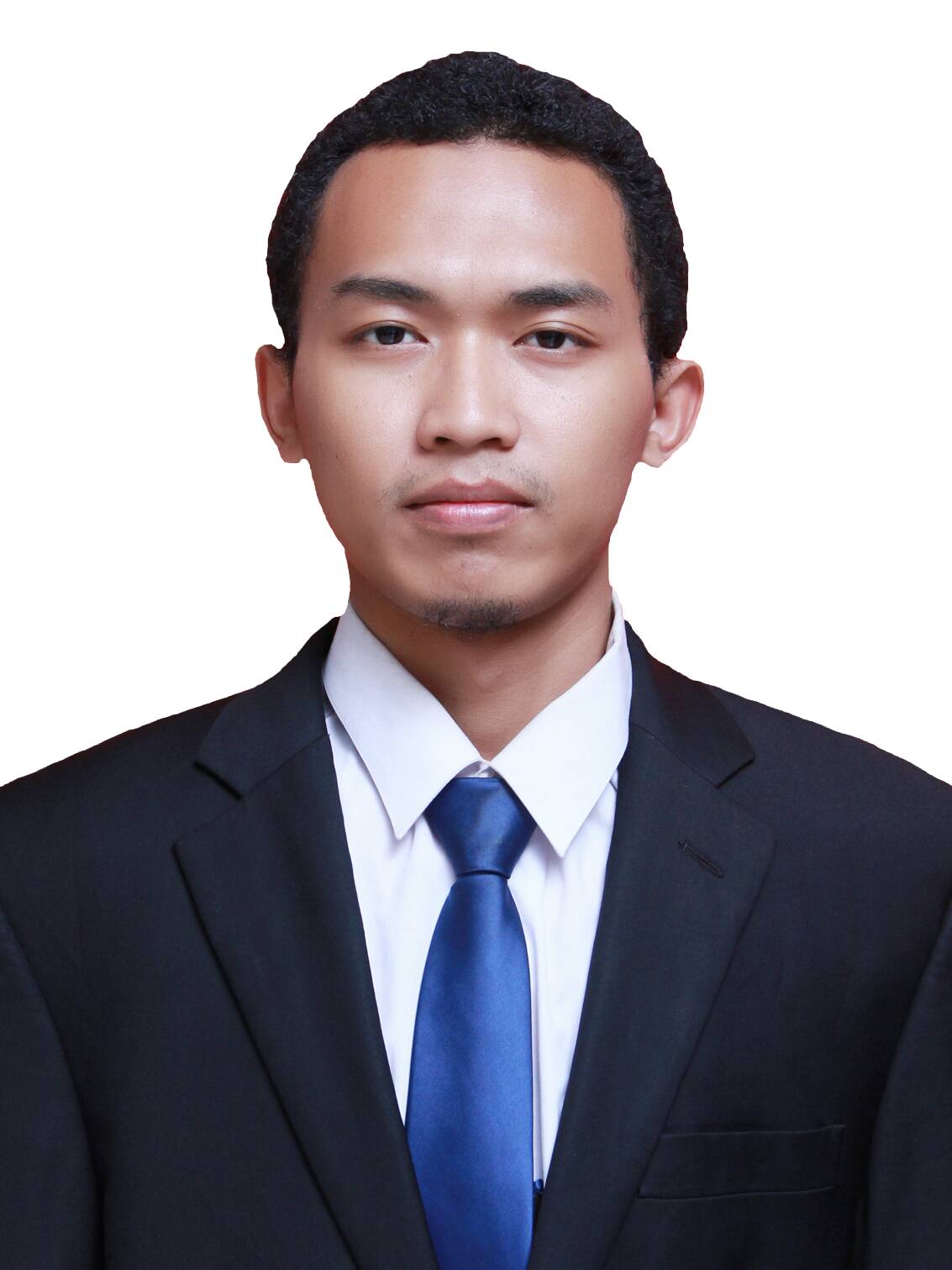}}]{Naufal Suryanto}
is a Postdoctoral Researcher specializing in AI Security at the IoT Research Center, Pusan National University, South Korea. His research focuses on advancing robust and trustworthy AI through practical adversarial machine learning and emerging technologies, with applications in computer vision and cybersecurity. Naufal has been recognized as a first author at top-tier conferences such as CVPR and ICCV, where he also received an Outstanding Reviewer award. His current work includes developing Generative AI for cybersecurity applications and researching adversarial attacks and defense systems. He earned his Ph.D. in Computer Engineering from Pusan National University in 2024 and his Bachelor’s degree in the same field from the Electronics Engineering Polytechnic Institute of Surabaya, Indonesia, in 2018. He is an active contributor to advancements in AI and cybersecurity.
\end{IEEEbiography}

\begin{IEEEbiography}[{\includegraphics[width=1in,height=1.25in,clip,keepaspectratio]{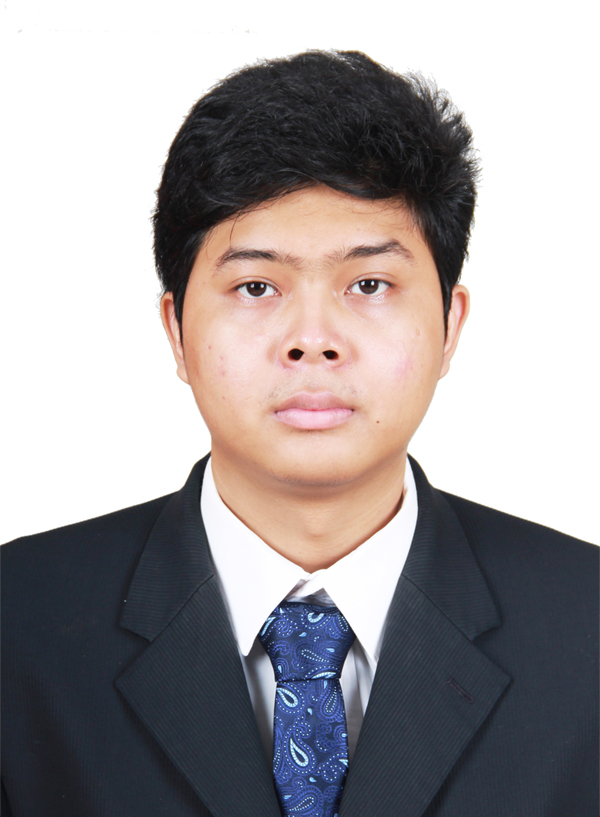}}]{Andro Aprila Adiputra}
is a Master researcher specialized in Artificial Intelligence and its application on Large Language Model(LLM) and Computer Vision. Obtained Bachelor's degree from Electronics Engineering Polytechnics Institute of Surabaya, Indonesia (2018 - 2022) and currently enrolling as Integrated Master and PhD student in Pusan National University, South Korea (2023 - present). His present research works are ranging from AI implementation for Security, Computer Vision for UAV Detection, and Pattern recognition.
\end{IEEEbiography}

\begin{IEEEbiography}[{\includegraphics[width=1in,height=1.25in,clip,keepaspectratio]{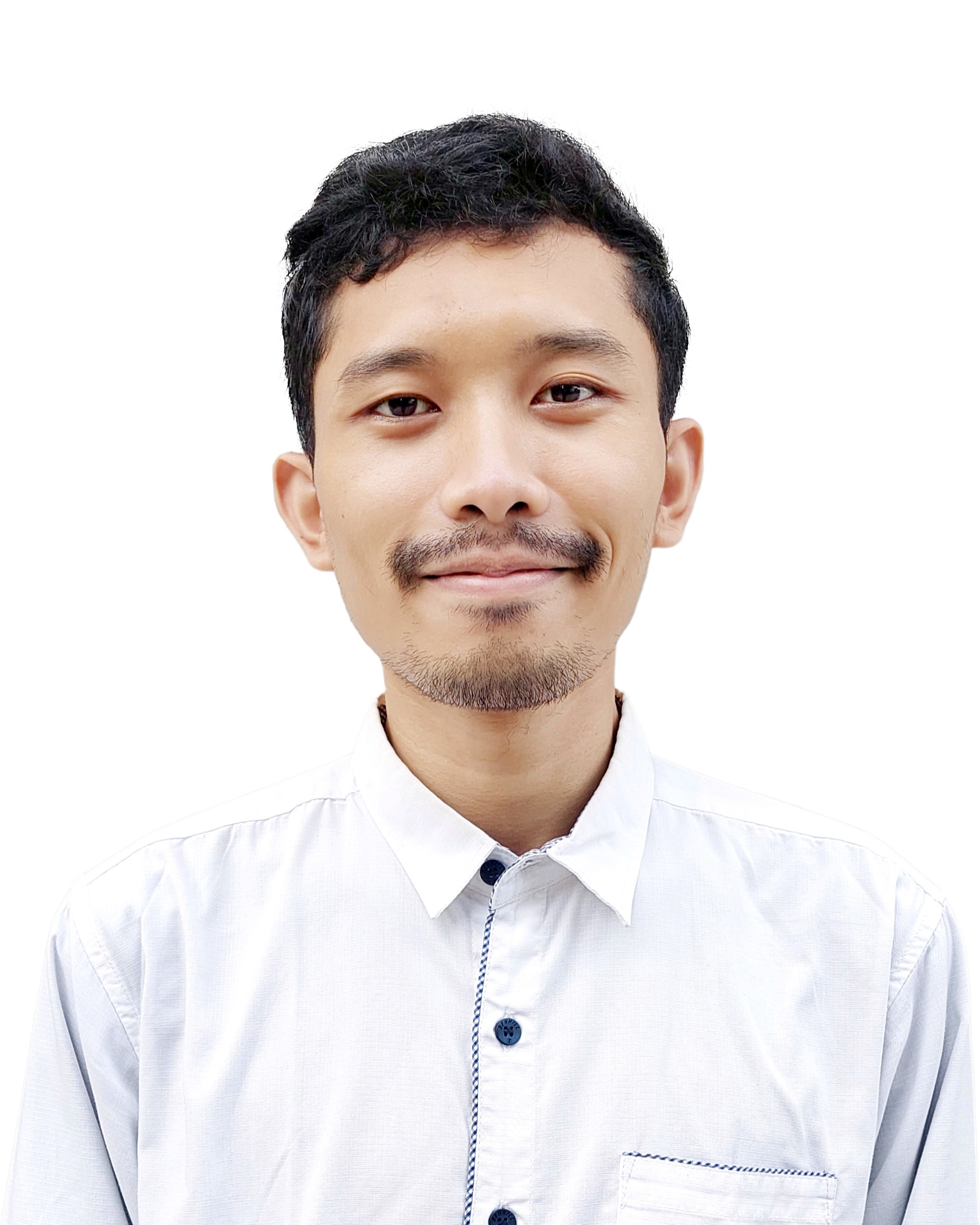}}]{Ahmada Yusril Kadiptya} received the B.Asc. degree in informatics engineering from the Electronic Engineering Polytechnic Institute of Surabaya, Indonesia, in 2019. He currently pursuing an Integrated Master and PhD degree in computer engineering at Pusan National University, South Korea (2023 - present). His current research interests including cybersecurity, system observability, and artificial intelligence.
\end{IEEEbiography}

\begin{IEEEbiography}[{\includegraphics[width=1in,height=1.25in,clip,keepaspectratio]{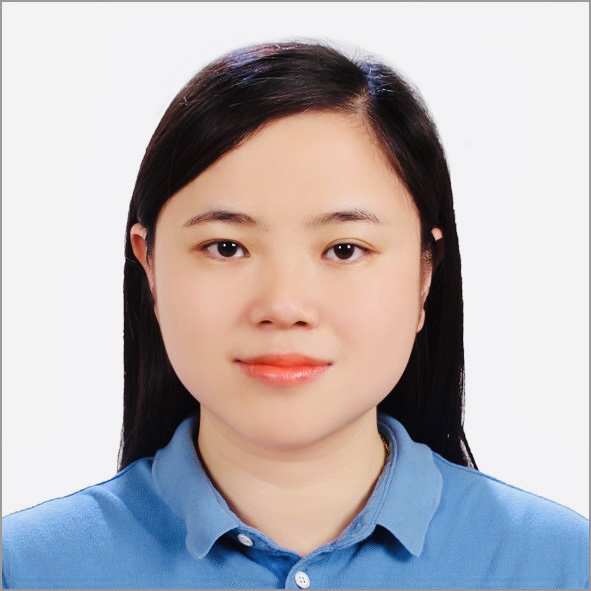}}]{THI-THU-HUONG LE} received the bachelor’s degree from Hung Yen University of Technology and Education (HYUTE), Vietnam, in 2007, the master’s degree from Hanoi University of Science and Technology (HUST), in 2013, and the Ph.D. degree from Pusan National University (PNU), South Korea, in 2020. She has three years of experience as a Postdoctoral Researcher with PNU, starting in 2020. She has seven years of experience as a Lecturer with HYUTE. She is a research professor at the Blockchain Platform Research Center, PNU. She has participated in machine learning projects like NILM, IDS, AI industry 4.0, AI security, and deep learning-based CFD. Her research interests include machine learning, deep learning, generative AI, data analysis, explainable AI, and signal processing. She has served several academic services, such as a guest editor, a technical committee, and a reviewer.
\end{IEEEbiography}

\begin{IEEEbiography}[{\includegraphics[width=1in,height=1.25in,clip,keepaspectratio]{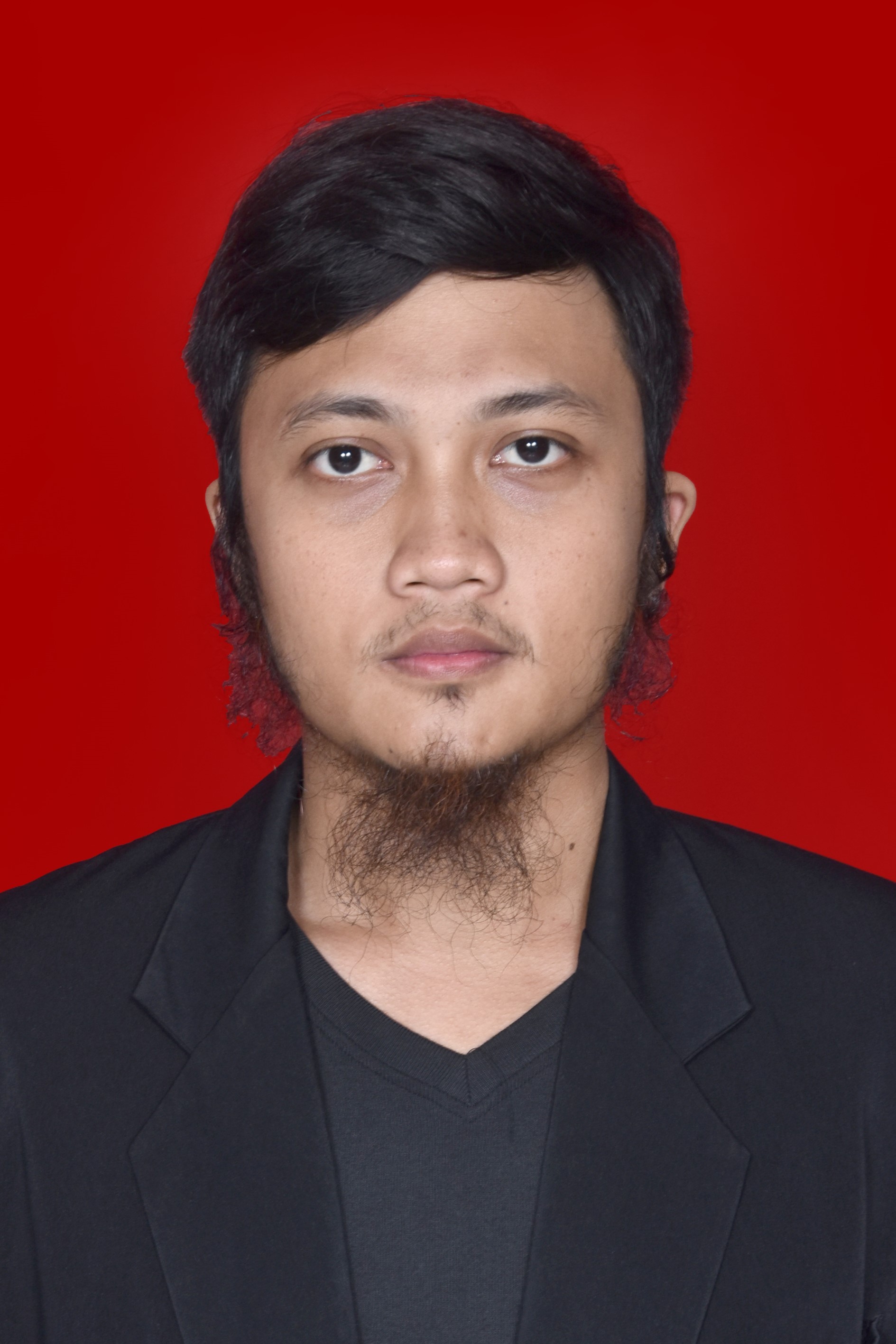}}]{DERRY PRATAMA} is a PhD researcher specializing in artificial intelligence, cybersecurity, and robotics, focusing on developing intelligent systems and methodologies to tackle complex challenges in these fields. His work encompasses AI-driven penetration testing, cryptographic optimization, and advanced robotic automation, with a keen interest in the interaction between large language models and automated systems to explore new paradigms in reverse engineering and hardware security. He earned a Bachelor's degree from the Sepuluh Nopember Institute of Technology, Indonesia (2011-2015), and a Master's degree from Pusan National University, South Korea (2019-2023), focusing on hardware security and UAV communication. Currently pursuing a Doctoral degree at Pusan National University (2023-present), his ongoing research interests include AI-driven cybersecurity validation, large language model augmentation for penetration testing, and the application of AI in autonomous systems.
\end{IEEEbiography}

\begin{IEEEbiography}[{\includegraphics[width=1in,height=1.25in,clip,keepaspectratio]{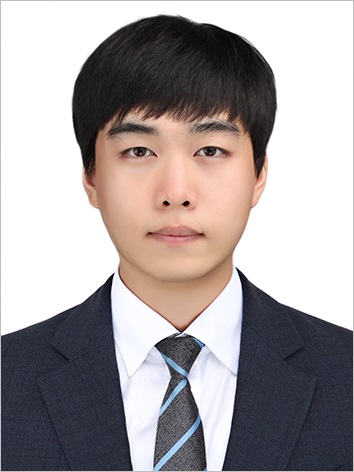}}]{YONGSU KIM} received the B.S. and M.S. degrees in computer engineering from Pusan National University (PNU), Busan, South Korea, and the Ph.D. degree in 2024. Since 2020, he has been with SmartM2M. His research interests include AI security, generative AI, and large language models.
\end{IEEEbiography}

\begin{IEEEbiography}[{\includegraphics[width=1in,height=1.25in,clip,keepaspectratio]{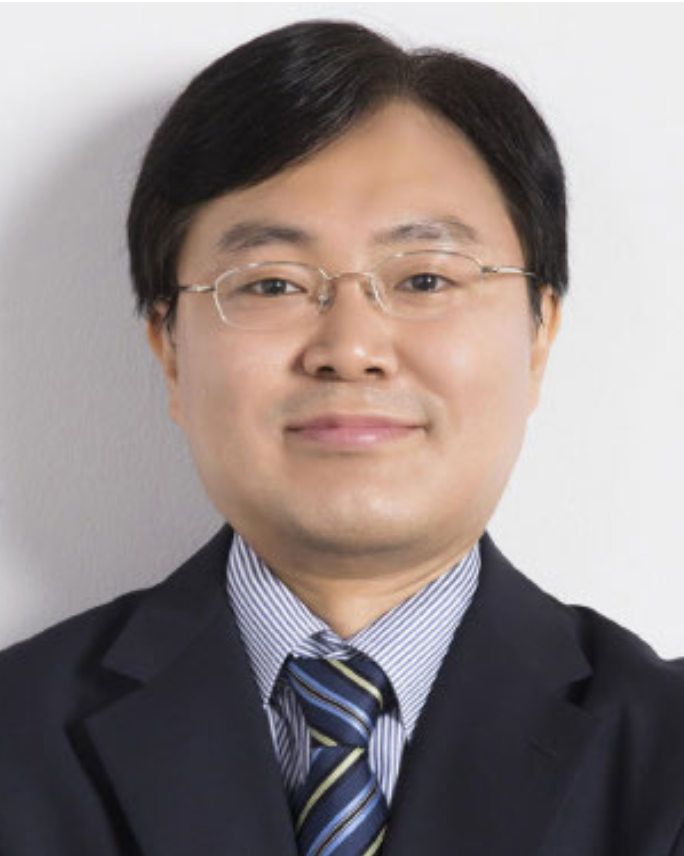}}]{HOWON KIM} (Member, IEEE) received the bachelor’s degree from Kyungpook National University (KNU), and the Ph.D. degree from Pohang University of Science and Technology (POSTECH). He is currently a Professor at the Department of Computer Science and Engineering, the Chief of Energy Internet of Things (IoT) with the IT Research Center (ITRC), and was the Chief of the Information Security Education Center (ISEC), Pusan National University (PNU). Before joining PNU, he was with the Electronics and Telecommunications Research Institute (ETRI) as a Team Leader for ten years beginning, in December 1998. He was a Visiting Postdoctoral Researcher with the Communication Security Group (COSY), Ruhr-University Bochum, Germany, from July 2003 to June 2004.
\end{IEEEbiography}

\EOD

\end{document}